\documentclass[11pt]{article}
\usepackage[preprint]{acl}
\usepackage{microtype}
\usepackage{fontspec}
\usepackage{amsmath,amssymb,unicode-math}
\usepackage{graphicx,booktabs,array,tabularx}
\usepackage{algorithm}
\usepackage{algpseudocode}
\algrenewcommand\algorithmicrequire{\textbf{Input:}}
\algnewcommand{\IfThen}[2]{\State \algorithmicif\ #1\ \algorithmicthen\ #2}
\makeatletter
\algnewcommand{\LongState}[1]{\State\parbox[t]{\dimexpr\linewidth-\ALG@thistlm-1pt\relax}{\raggedright\strut#1\strut}}
\makeatother
\usepackage{xurl}
\usepackage{placeins}  
\providecommand{\tightlist}{\setlength{\itemsep}{0pt}\setlength{\parskip}{0pt}}
\hypersetup{pdftitle={SAG2: Embedding-Free Multi-Hop Retrieval with BM25 and Decision Models}, pdfauthor={Yuchao Wu, XingCheng Liang, Guanxian Li}}

\title{SAG2: Embedding-Free Multi-Hop Retrieval with BM25 and Decision Models}
\author{Yuchao Wu\footnotemark, XingCheng Liang, Guanxian Li \\
  Zleap AI \\
  \texttt{jomy@zleap.com, lensen@zleap.com, li\_guanxian@foxmail.com}}

\begin{document}
\maketitle
{\renewcommand{\thefootnote}{\textasteriskcentered}%
 \footnotetext{Correspondence to: \href{mailto:jomy@zleap.com}{\texttt{jomy@zleap.com}}.}}

\begin{abstract}
Multi-hop retrieval must find passages that become useful only after intermediate evidence is retrieved. We present SAG2, an embedding-free retriever whose only persistent structure is a SQLite lexical index, built without an LLM. Where SAG and HippoRAG 2 rely on a generative LLM for graph extraction and selection, SAG2 uses a decision model. Given the question and a shared context of selected retrieved passages, the decision model scores whether each candidate supplies an answer-chain fact. Terms activated from the top-scoring passages retrieve unvisited passages through the lexical index, which holds a passage--term hypergraph implicitly. SAG2 reaches Recall@10 of 86.62\%, 97.43\%, and 99.00\% on 1,000-question subsets of MuSiQue, 2WikiMultiHopQA, and HotpotQA, the highest among the compared systems. With a decision model priced at about a fifth of SAG's LLM, indexing and retrieval on MuSiQue cost \$4.33 against \$14.85 for SAG. On EnterpriseRAG-Bench, SAG2 reaches 81.12\% document Recall@10. As the benchmark's nested corpus grows 447.5-fold, SAG2's Recall@5 lead over BM25 alone widens from 7.28 to 14.76 points, at a nearly constant model cost per question. Because indexing makes no model call, SAG2 suits corpora that change often or must be searchable as soon as they are indexed.\footnote{Benchmark and evaluation code:\newline
  {\urlstyle{tt}\url{https://github.com/Zleap-AI/SAG-Benchmark}}}
\end{abstract}

\begin{figure}[t]
\centering
\includegraphics[width=\columnwidth]{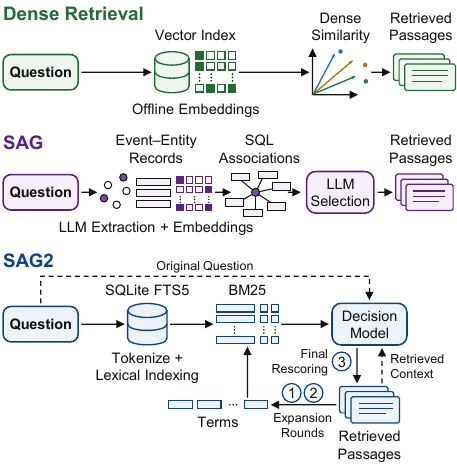}
\caption{Retrieval architectures. Dense retrieval precomputes passage embeddings. SAG stores an LLM-extracted event--entity hypergraph together with dense representations and activates shared-entity associations at query time. SAG2 stores original text and lexical occurrences in SQLite and builds associations online with BM25 and a decision model. Solid arrows denote retrieval or expansion; dashed arrows pass the question and retrieved context to the decision model.}\label{fig:arch}
\end{figure}

\section{Introduction}
Multi-hop questions require evidence distributed across passages, and the route to a supporting passage may emerge only after an intermediate fact is retrieved \citep{trivedi2022musique}. Consider the MuSiQue question ``What county is Erik Hort's birthplace a part of?'', which \citet{gutierrez2025hipporag2} also use as an example. A system must first find that Hort was born in Montebello, New York. The passage that places Montebello in Rockland County shares that place name with the first passage but matches few words of the question. Iterative retrievers follow such links through the entities in retrieved passages \citep{ding2019cogqa,das2019multistep}. Yet each passage contains many terms that could serve as bridges, and retrieval must decide which to follow as evidence accumulates.

Dense retrieval embeds every passage in advance \citep{karpukhin2020dpr}, and structure-augmented retrieval also runs an LLM over every passage. SAG extracts an event--entity hypergraph and activates its shared-entity associations at query time, alongside dense retrieval and model selection \citep{wu2026sag}; HippoRAG 2 links passages through a knowledge graph of LLM-extracted triples \citep{gutierrez2025hipporag2}. Indexing the 11,656 MuSiQue passages takes HippoRAG 2 and SAG 2.30 and 4.57 hours (Table~\ref{tab:efficiency}), and every document added later must pass through the same extraction. Feedback methods expand from the results of a query through relevance models \citep{lavrenko2001relevance}, neurally selected feedback text \citep{zheng2020bertqe}, or a precomputed corpus graph \citep{macavaney2022gar}. They expand toward the target of the original query, whereas multi-hop retrieval needs terms that lead to a different target at each step \citep{xiong2021mdr}. We therefore ask whether the evidence judged so far can decide which lexical connections to follow, so that an index built without embeddings or an LLM reaches multi-hop evidence.

We let each candidate's contribution to the answer chain decide which connections to follow. A decision model (DM) scores one candidate at a time for a given question, context, and instruction, much as LLMs have judged the relevance or utility of passages \citep{thomas2024llmpreferences,zhang2024utility}. Later rounds share selected retrieved passages as context, which establishes how a candidate's fact connects to the answer chain.

We implement this mechanism in \textbf{SAG2} as a \textbf{local hypergraph with dynamic term activation} (Figure~\ref{fig:arch}). A lexical inverted index holds a passage--term hypergraph implicitly, each passage a hyperedge over its terms. For each question, SAG2 computes only the part of this hypergraph that retrieval explores. The BM25 candidates that the decision model ranks highest become weighted seeds, terms activated from their sentences retrieve unvisited passages, and the decision model judges these passages in the next round (Section~\ref{sec:hypergraph}).

We measure evidence recall mainly by Recall@10 (R@10) and make the following contributions:
\begin{itemize}
\tightlist
\item We formulate multi-hop retrieval over a lexical index as a local hypergraph with dynamic term activation: an inverted index holds a passage--term hypergraph implicitly, and for each question retrieval computes only the part reached by terms activated from decision-model-judged seeds. This local hypergraph reaches multi-hop evidence beyond the candidate pool.
\item We propose SAG2, an embedding-free retriever that replaces the generative LLM that SAG and HippoRAG~2 use for graph extraction and selection with a query-time decision model, so its index is built without an LLM.
\item On MuSiQue, 2WikiMultiHopQA, and HotpotQA, SAG2 reaches the highest R@10 among the compared systems. Controlled comparisons attribute its gain over evidence scoring of the candidate pool to term expansion and the shared context, and its lead over SAG holds with open-weight decision models of 9B and 35B parameters. On EnterpriseRAG-Bench \citep{sun2026enterpriserag}, its lead over BM25 widens as the corpus grows, at a nearly constant model cost per question.
\end{itemize}

\section{Related Work}\label{sec:related}
\textbf{Reranking, judgment, and feedback.} Rerankers reorder a retrieved list, which bounds their recall \citep{nogueira2020monot5,sun2023rankgpt,zhuang2024setwise}. LLMs also judge whether a passage is relevant or useful \citep{thomas2024llmpreferences,zhang2024utility}, and Jev has served as a judge and reranker of fixed BM25 candidates \citep{fernandes2026jevjudge}. Feedback methods expand from initial results: relevance models and their LLM-filtered variants \citep{lavrenko2001relevance,abduljaleel2004rm3,otero2026llmprf}, neural or LLM selection of feedback text \citep{zheng2020bertqe,lei2024csqe}, and GAR, which follows a precomputed corpus graph from highly reranked documents \citep{macavaney2022gar,rathee2025slidegar}. SAG2 moves the judgment into the retrieval loop. One answer-chain judgment in a shared context both selects the feedback passages and scales the expansion priorities of their terms; each activated term then retrieves its own list of unvisited passages, and candidates are rescored over several rounds.

\textbf{Graph- and hypergraph-based retrieval.} Lexical statistics define many text graphs: TF--IDF supplies the word--sentence edge features of HeterSumGraph \citep{wang2020hetersumgraph}, TF--IDF or BM25 selects nodes of keyphrase-generation graphs \citep{ye2021keyphrasegraph,kim2021structurekeyphrase}, and a TF--IDF-style score reselects the keywords of a co-occurrence graph in each round \citep{zhang2021keywordgraph}. In retrieval, OkaSim ranks BM25-filtered documents by SimRank over a document--term graph \citep{champclaux2009okasim}, and GRADA and GAR link retrieved and corpus documents, respectively, by BM25 or other similarities \citep{zheng2025grada,macavaney2022gar}. Hypergraph models treat each document as a hyperedge over its terms, held implicitly or materialized \citep{devezas2019hypergraphofentity,spitz2020hypergraph}, and Graph-of-word stores only graph-derived term weights in the inverted index \citep{rousseau2013graphofword}. Many graph-based RAG systems instead use an LLM to extract triples \citep{gutierrez2024hipporag,gutierrez2025hipporag2}, n-ary facts or entities \citep{luo2025hypergraphrag,feng2026hyperrag,wang2026hgrag,an2026hyperprove}, passage-linking pseudo-queries \citep{liu2025hoprag}, or event--entity records \citep{wu2026sag}. Others replace most LLM extraction with co-occurrence, keywords, or recognized entities but usually compute embeddings offline \citep{edge2024lazygraphrag,bonifazi2026tigrag,zhuang2025linearrag,song2026ehrag,huang2025ketrag,liu2026hypersu}. SAG2 writes no edges or embeddings at indexing. For each question, decision-model judgments select the seed passages whose activated terms retrieve passages outside the candidate pool.

\textbf{Multi-hop iterative and agentic retrieval.} Multi-hop retrievers condition each step on earlier evidence through extracted entities \citep{ding2019cogqa,das2019multistep}, trained encoders or condensers \citep{xiong2021mdr,zhang2024beamretrieval,khattab2021baleen}, or generated queries, reasoning, or LLM judgments \citep{qi2019golden,trivedi2023ircot,wang2025elite}; LLM agents search while they reason \citep{yao2023react,jin2025searchr1,li2026dci}. SAG2 instead selects next-round terms from passages the decision model has judged and retrieves with them lexically, without generating query text, training a reader or encoder, or computing embeddings. Appendix~\ref{app:related} extends this section.

\section{SAG2}
\begin{figure*}[t]
\centering
\includegraphics[width=\textwidth,trim=0 56 314 0,clip]{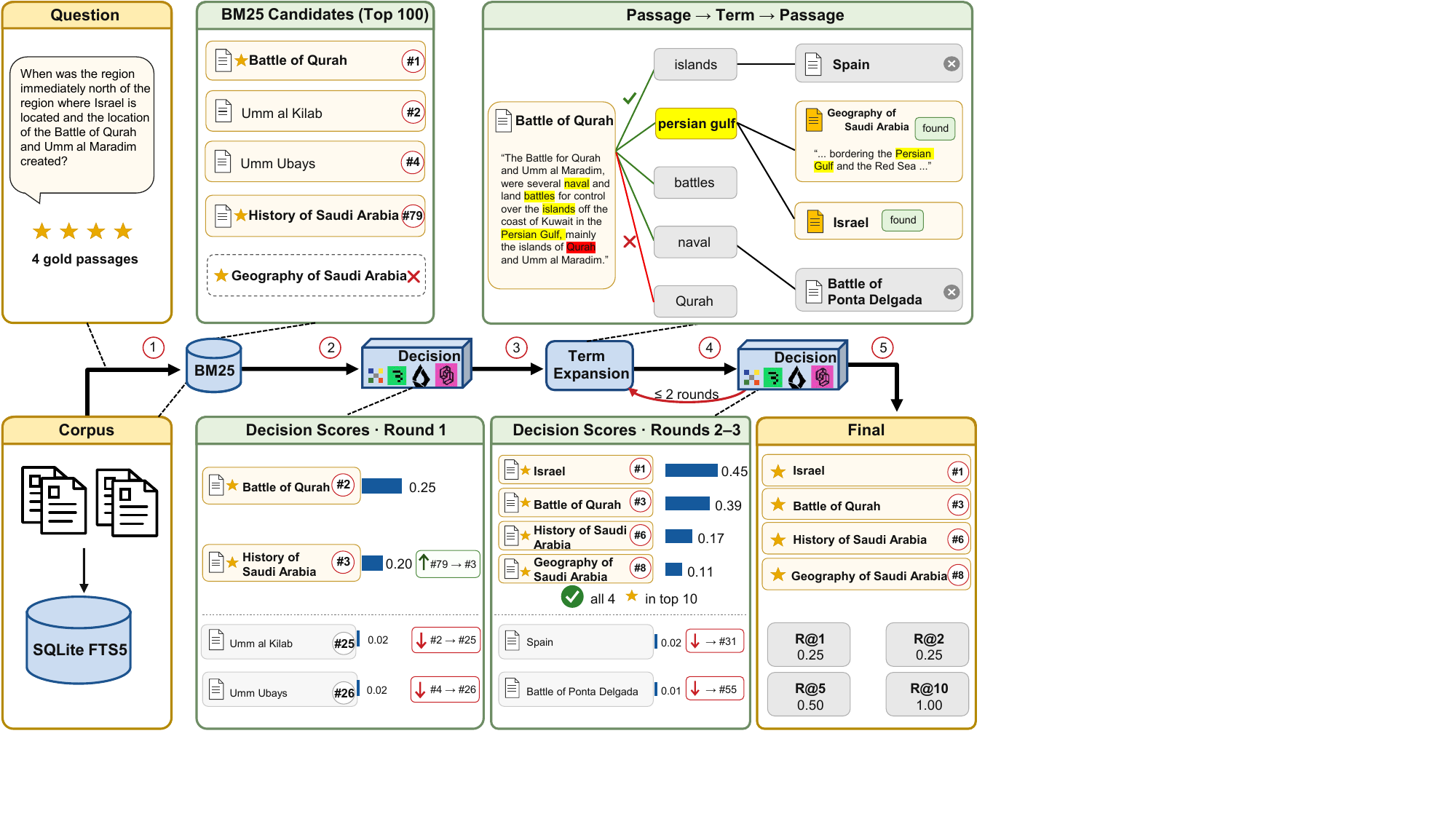}
\caption{Retrieval trace of SAG2 for one MuSiQue question with four supporting passages (stars). (1) BM25 ranks History of Saudi Arabia outside its top ten and Geography of Saudi Arabia outside its top 100. (2) Judged by the decision model against the question alone, History of Saudi Arabia rises into the top ten. (3) Terms activated in the top-ranked seeds, such as \textit{persian gulf} from Battle of Qurah, retrieve unvisited passages. (4) The decision model rescores the enlarged pool with the retrieved passages as shared context. (5) All four supporting passages reach the final top ten, Geography of Saudi Arabia through the bridge term \textit{persian gulf}.}\label{fig:workflow}
\end{figure*}

SAG2 proceeds in five stages (Algorithm~\ref{alg:sag2}): \textbf{lexical retrieval} takes the top \(B=100\) BM25 matches of the question as the candidate pool; \textbf{evidence scoring} has the decision model judge each candidate in a shared context; \textbf{term activation} selects terms from the highest-scoring seeds; \textbf{term expansion} retrieves unvisited passages for each activated term; and after at most \(H=2\) expansion rounds, \textbf{final rescoring} reassesses the pool. Figure~\ref{fig:workflow} follows one MuSiQue question through these stages. Each hop follows a lexical bridge: a next-hop passage is reached through a token, a name phrase, or a title-derived alias that it shares with a passage the decision model has judged to be evidence (Section~\ref{sec:hypergraph}).

\subsection{Local hypergraph with dynamic term activation}\label{sec:hypergraph}
Let \(\mathcal D\) be the set of indexed units, which we call passages, and \(\mathcal V\) the set of terms, namely normalized tokens and multiword name phrases (Section~\ref{sec:indexing}). The inverted index determines an occurrence relation \(M(d,t)\in\{0,1\}\): \(M(d,t)=1\) when the tokens of \(t\) occur consecutively in the title, heading, or body field of \(d\), which is what the FTS5 phrase query for \(t\) tests. Each passage thus defines a hyperedge \(e(d)=\{t\in\mathcal V:M(d,t)=1\}\), and the corpus forms the passage--term hypergraph \(\mathcal G=(\mathcal V,\{e(d)\}_{d\in\mathcal D})\). The index holds \(\mathcal G\) implicitly \citep{spitz2020hypergraph}: the postings and token positions that BM25 already uses are the term-side adjacency lists of \(M\). Indexing therefore writes no hyperedge, edge, or adjacency list beyond the FTS5 index, the corpus statistics, and the title-derived name aliases.

Term activation uses a finer, sentence-level relation. Let \(\operatorname{sent}(d)\) be the body sentences of \(d\), and let the local frequency \(\operatorname{tf}(t,d)\) count the occurrences of \(t\) in them, where a matched multiword name counts once and its component words are not counted again. A passage can be reached through any term in \(e(d)\), but it activates only the terms with \(\operatorname{tf}(t,d)>0\).

For a question \(q\), let \(\Omega_r\) be the passages visited up to expansion round \(r\), including the candidate pool \(C_0\), and \(T_r=T_{r-1}\cup X_r\) the terms activated so far, where \(X_r\) is the set activated in round \(r\) and \(T_0=\varnothing\). The local hypergraph is the part of \(\mathcal G\) that the question has reached,
\begin{equation}
\mathcal G_r(q)=\bigl(T_r,\ \{e(d)\cap T_r\}_{d\in\Omega_r}\bigr).\label{eq:local}
\end{equation}
Evidence scores \(s_r(d)\in[0,1]\) from the decision model (Section~\ref{sec:indexing}) select the seeds from which \(\mathcal G_r(q)\) grows. After scoring round \(r\), the seeds \(S_r\subseteq\Omega_r\) are up to \(K=10\) of the highest-scoring unexpanded passages, and only their terms can be activated. Seeds are chosen by judgment rather than by BM25 rank. A judged seed holds a fact of the answer chain, so its terms lead to the next hop, whereas BM25 rank measures only overlap with the question. An expansion round takes two steps. Term activation, the hyperedge-to-node step, ranks terms by their expansion priority, scoring each term through its strongest seed:
\begin{equation}
a_r(t)=\max_{d\in S_r}\,w_r(d)\,\phi_q(t,d),\label{eq:activation}
\end{equation}
with the seed weight, the activation weight, and the question connection
\begin{align}
w_r(d)&=0.5+0.5\,\frac{s_r(d)}{\max_{d'\in S_r}s_r(d')},\notag\\
\phi_q(t,d)&=\bigl(1+\log\operatorname{tf}(t,d)\bigr)\,I(t)\,\bigl(1+g_q(t,d)\bigr),\notag\\
g_q(t,d)&=\max_{u\in\operatorname{sent}(d),\,t\in u}\;\sum_{v\in\operatorname{tok}(u)\cap\operatorname{tok}(q)}I(v),\label{eq:weights}
\end{align}
where \(w_r(d)\in[0.5,1]\) downweights lower-scoring seeds and equals \(0.5\) when every seed has \(s_r(d)=0\). The activation weight \(\phi_q(t,d)\) is zero when \(\operatorname{tf}(t,d)=0\). Logarithms are natural, and \(I(t)\) is the BM25 inverse document frequency of \(t\) over passage bodies \citep{robertson2009bm25}. The question connection \(g_q(t,d)\) is the largest summed IDF of the question tokens that a sentence of \(d\) containing \(t\) shares with \(q\), where \(\operatorname{tok}(\cdot)\) is the set of distinct normalized tokens of a text, including the words within names. Expansion priority thus favors terms frequent in a high-scoring seed, rare in the corpus, and located in sentences that share tokens with the question. Terms consisting only of question tokens or with zero body-level document frequency are ineligible, and the \(L=20\) eligible terms of highest priority form \(X_{r+1}\).

Term expansion, the node-to-hyperedge step, follows \(M\) from each activated term \(t\in X_{r+1}\) to passages not yet visited,
\begin{equation}
C_{r+1}(t)=\{d\in\mathcal D\setminus\Omega_r:M(d,t)=1\}.\label{eq:retrieval}
\end{equation}
A new passage must contain \(t\) but need not match any question token. Each term's list is ranked by BM25 over the OR expression of \(X_{r+1}\) and the question tokens. Terms retrieve separate lists because a next-hop passage may share only one bridge term with its seed: it ranks high in that term's list but behind multi-term matches in a single OR query. The lists are interleaved rank by rank in order of expansion priority (Eq.~\ref{eq:activation}), without duplicates, into at most \(B=100\) new candidates \(C_{r+1}\), so that every activated term receives a share of the slots. The decision model then rescores the pool containing them, which yields \(s_{r+1}\) and the next seeds. Question tokens are never recorded in \(T_r\). An earlier term stays in \(T_{r+1}\) but is queried again only if it is reactivated. The corpus text thus fixes \(M\) and \(\operatorname{tf}\), whereas judgments decide which hyperedges become seeds and which passages enter \(\mathcal G_{r+1}(q)\) (Eq.~\ref{eq:local}).

\subsection{Indexing, evidence scores, and candidate pools}\label{sec:indexing}
\textbf{Indexing.} SQLite FTS5 \citep{sqlite2026fts5} indexes the title, heading, and body of each passage with BM25 and stores the original text, the number of indexed units, and the body-level document frequency of every token. Multiword name phrases are recognized by a capitalization rule when terms are extracted from seed sentences. Source texts are split into units of about 1,000 characters, at most 1,200, and each of the 11,656 MuSiQue passages forms a single unit. Appendix~\ref{app:params} gives tokenization, chunking, and name handling.

\textbf{Lexical retrieval and evidence scores.} Lexical retrieval takes the top \(B=100\) BM25 matches for the tokenized question as the candidate pool \(C_0\). The evidence score of candidate \(d\) is \(s_r(d)=\mathrm{DM}(q,d;E_r)\in[0,1]\), the judgment of the decision model (Jev by default; Section~\ref{sec:stats}) given the shared context \(E_r\) of round \(r\). The score enters the seed weight of Eq.~\ref{eq:weights}. The answer-chain instruction (Appendix~\ref{app:params}) asks whether the candidate itself supplies a necessary fact, including an intermediate identification fact, in a fully supported answer chain while respecting entity identities and relation constraints. The context can link the candidate to the chain, but facts found only in the context, topical association, and optional background earn no credit. The context is shared because a next-hop passage states its fact about an intermediate entity, not about the question, so only the retrieved passage that introduces that entity ties the fact to the chain. The context consists of retrieved text and model outputs: it is empty in the first scoring round, holds the retained seeds and the first \(K=10\) new candidates in the first expansion round, and holds the top ten passages of the previous ranking, with their tentative scores, in final rescoring.

\textbf{Candidate pools.} Each expansion round rescores its new candidates together with the seeds accumulated so far and the \(F=5\) top-ranked passages of lexical retrieval, which stay protected in every pool. Final rescoring keeps the size of its pool but admits at most 20 unvisited passages (the final-round incoming cap), filling the remaining slots with previously scored candidates. Earlier candidates are rescored because each was judged before the later hops were retrieved, so the final round can recognize a first-hop passage whose link to the chain runs through a later hop. Ties in evidence score are broken by the reciprocal rank of first retrieval \citep{cormack2009rrf}. The output keeps at most \(B=100\) passages, which evaluation maps to unique source passages.

\textbf{Bounded retrieval.} SAG2 performs at most \(H=2\) expansion rounds after lexical retrieval, giving at most three scoring rounds, and stops earlier when no usable lexical expression, unexpanded seed, expansion term, or new candidate remains (Algorithm~\ref{alg:sag2}; defaults in Table~\ref{tab:params}, Appendix~\ref{app:additional}). Because protected passages and accumulated seeds are rescored with the new candidates, a question receives at most 340 passage judgments.

\begin{algorithm}[t]
\caption{SAG2 retrieval. Round-\(r\) objects of Section~\ref{sec:hypergraph} drop the subscript (\(S=S_r\), \(\Omega=\Omega_r\), \(C=C_r\), \(E=E_r\)).}\label{alg:sag2}
\small
\begin{algorithmic}[1]
\Require question $q$, lexical index $\mathcal{D}$, decision model $\mathrm{DM}$; $B{=}100$, $K{=}10$, $L{=}20$, $F{=}5$, $H{=}2$
\State $P, A, \Omega, \mathit{Expanded}, \mathit{History}, \mathit{Ranking} \gets$ empty
\For{$r = 0,\dots,H$}
  \LongState{$C \gets \mathrm{BM25}(q, B)$ if $r = 0$, else $\mathrm{Expand}(q, S, \Omega, B, L)$ by Eqs.~\ref{eq:activation}--\ref{eq:retrieval} \Comment{lexical retrieval, then term expansion}}
  \IfThen{$C$ is empty}{\textbf{stop}}
  \IfThen{$r = 0$}{$P \gets$ first $F$ items of $C$}
  \State $A \gets \mathrm{unique}(A + S)$ if $r > 0$, else empty
  \State $U \gets C$ if $r = 0$, else $\mathrm{unique}(P + A + C)$
  \State $W \gets U$
  \State $E \gets$ empty if $r = 0$, else $\mathrm{unique}(A + \mathrm{first}_K(C))$
  \If{$r = H$} \Comment{final rescoring}
    \State $\mathit{New} \gets$ first 20 items of $U$ not in $\mathit{History}$
    \LongState{$\mathit{Old} \gets \mathrm{unique}$(already-scored items of $U$ + $\mathit{History}$ by latest score)}
    \State $W \gets$ first $(|U| - |\mathit{New}|)$ items of $\mathit{Old} + \mathit{New}$
    \LongState{$E \gets$ first 10 items of previous $\mathit{Ranking}$, with tentative prior scores}
  \EndIf
  \LongState{$\mathit{Ranking} \gets$ rank $W$ using contextual $\mathrm{DM}(q, d; E)$ and deterministic tie breakers \Comment{evidence scoring}}
  \LongState{$\mathit{History} \gets$ update $\mathit{History}$ with the latest scores in $\mathit{Ranking}$}
  \State $\Omega \gets \Omega \cup \mathrm{identifiers}(W)$
  \IfThen{$r = H$}{\textbf{stop}}
  \LongState{$S \gets$ first $K$ items of $\mathit{Ranking}$ whose identifiers are not in $\mathit{Expanded}$ \Comment{seeds for term activation}}
  \IfThen{$S$ is empty}{\textbf{stop}}
  \State $\mathit{Expanded} \gets \mathit{Expanded} \cup \mathrm{identifiers}(S)$
  \IfThen{no usable expansion term from $S$}{\textbf{stop}}
\EndFor
\State $\mathit{Ranking} \gets$ first $B$ items of $\mathit{Ranking}$
\LongState{\textbf{return} $\mathit{Ranking}$, mapped to original passages}
\end{algorithmic}
\end{algorithm}

\section{Experimental Setup}\label{sec:stats}
\textbf{Benchmarks and metrics.} We evaluate on the 1,000-question development subsets of MuSiQue \citep{trivedi2022musique}, 2WikiMultiHopQA \citep{ho2020twowiki}, and HotpotQA \citep{yang2018hotpotqa} and their retrieval corpora of 11,656, 6,119, and 9,811 passages, the setting of HippoRAG \citep{gutierrez2024hipporag} that HippoRAG~2 \citep{gutierrez2025hipporag2} and SAG \citep{wu2026sag} also use. On EnterpriseRAG-Bench \citep{sun2026enterpriserag} (EnterpriseRAG in tables), we measure document-level recall for the 470 questions with reference documents on its full corpus of 511,959 source documents, mapping retrieved chunks to unique source documents (Section~\ref{sec:enterprise}). We report Recall@k (R@k), the fraction of a question's supporting passages among the first \(k\) unique source passages, macro-averaged over questions. AllEvidence@k (AE@k), the AR@k of HippoRAG \citep{gutierrez2024hipporag}, is the fraction of questions whose supporting passages all appear among the first \(k\).

\begin{table*}[t]
\centering
\setlength{\tabcolsep}{3.2pt}
\renewcommand{\arraystretch}{1.12}
\resizebox{\textwidth}{!}{%
\begin{tabular}{@{}l*{12}{r}@{}}
\toprule
 & \multicolumn{4}{c}{MuSiQue} & \multicolumn{4}{c}{2WikiMultiHopQA} & \multicolumn{4}{c}{HotpotQA} \\
\cmidrule(lr){2-5}\cmidrule(lr){6-9}\cmidrule(lr){10-13}
Method & R@1 & R@2 & R@5 & R@10 & R@1 & R@2 & R@5 & R@10 & R@1 & R@2 & R@5 & R@10 \\
\midrule
\multicolumn{13}{@{}l}{\itshape Simple Baselines} \\
BM25 & 28.71 & 37.50 & 47.66 & 54.50 & 40.25 & 58.18 & 67.08 & 71.58 & 39.75 & 58.05 & 74.45 & 86.95 \\
Contriever & 24.50 & 33.54 & 45.32 & 53.58 & 34.40 & 48.60 & 59.25 & 65.20 & 39.30 & 57.45 & 74.85 & 82.75 \\
\multicolumn{13}{@{}l}{\itshape Large Embedding Models} \\
BGE-Large-EN-v1.5 & 29.42 & 41.68 & 56.32 & 64.08 & 41.48 & 62.28 & 69.83 & 73.75 & 45.45 & 76.20 & 89.20 & 93.35 \\
GTE-Qwen2-7B & \underline{35.19} & 48.06 & 61.92 & 70.80 & 43.45 & 65.58 & 74.18 & 78.63 & 46.40 & 73.95 & 88.45 & 93.55 \\
GritLM-7B & 34.86 & 50.70 & 66.22 & 75.19 & 41.85 & 66.23 & 75.62 & 79.40 & 46.40 & 81.15 & 93.40 & 97.20 \\
NV-Embed-v2 & 34.02 & 51.08 & 68.84 & 77.27 & 43.23 & 68.93 & 77.00 & 80.75 & 46.90 & 85.55 & 95.25 & \underline{97.70} \\
\multicolumn{13}{@{}l}{\itshape Structure-Augmented RAG} \\
HippoRAG 2 & 30.65 & 49.52 & 65.13 & 73.76 & 42.38 & 76.55 & 90.35 & 93.40 & 44.40 & 78.35 & 94.35 & 97.15 \\
SAG & \textbf{36.82} & \textbf{63.62} & \underline{80.36} & \underline{83.37} & \textbf{43.81} & \underline{83.93} & \underline{93.34} & \underline{93.57} & \underline{47.80} & \underline{91.55} & \underline{96.50} & \underline{97.70} \\
\midrule
SAG2 (Ours) & 34.30 & \underline{62.18} & \textbf{80.68} & \textbf{86.62} & \underline{43.55} & \textbf{85.50} & \textbf{97.35} & \textbf{97.43} & \textbf{48.30} & \textbf{93.60} & \textbf{98.45} & \textbf{99.00} \\
\bottomrule
\end{tabular}%
}
\caption{Passage recall (\%) on the 1,000-question subsets of three multi-hop benchmarks. BM25 is the lexical retrieval of SAG2; all other baseline results are as reported by \citet{wu2026sag}. Bold and underline mark the best and second-best values in each column.}\label{tab:main-results}
\end{table*}

\textbf{Baselines.} Table~\ref{tab:main-results} compares SAG2 with BM25~\citep{robertson2009bm25} (its own lexical retrieval), the dense retrievers Contriever~\citep{izacard2022contriever}, BGE-Large-EN-v1.5~\citep{xiao2024cpack}, GTE-Qwen2-7B~\citep{li2023gte}, GritLM-7B~\citep{muennighoff2024gritlm}, and NV-Embed-v2~\citep{lee2025nvembed}, HippoRAG 2~\citep{gutierrez2025hipporag2}, and SAG~\citep{wu2026sag}. Results for the dense and structure-augmented systems are from \citet{wu2026sag}; there, the structure-augmented systems use Qwen3.6-Flash as the reader, extractor, and selector and BGE-Large-EN-v1.5 as the retriever.

\textbf{Implementation.} SAG2 is implemented in Python on SQLite FTS5. The default decision model is Jev (\texttt{jev-1.14}), a System One model from TypeSafe AI that outputs probabilistic structured decisions instead of generating text \citep{almeida2026jev}. Through TypeSafe AI's Noul interface \citep{typesafe2026noul,typesafe2026models}, SAG2 uses the returned probability as \(s_r(d)\). A passage judgment is the decision model's evaluation of one candidate for a given question and context, and rescoring a retained passage counts as a new judgment. Agentic SAG2 adds a generative model, gpt-6.1-sol with high reasoning effort, that calls SAG2 as a retrieval tool at most three times per question. Each call runs one expansion and returns 20 passages, and the model ranks ten final passages from the returned candidates (Appendix~\ref{app:agentic}). We assess significance with paired tests over questions, mainly the percentile bootstrap \citep{efron1993bootstrap,koehn2004significance}, and mark it in the tables.

\section{Results and Analysis}
\subsection{Main retrieval results}
SAG2 reaches the highest R@10 on all three datasets (Table~\ref{tab:main-results}), and its advantage lies in bringing the later-hop passages into the top ten. On MuSiQue its 86.62\% lies 3.25 points above SAG; on 2WikiMultiHopQA and HotpotQA it exceeds SAG by 3.86 and 1.30 points. SAG2 also leads at R@5 on all three, and its margin over SAG is larger at R@5 and R@10 than at R@1 on every dataset, largest on MuSiQue and 2WikiMultiHopQA. SAG2 and BM25 share one index, so their gap comes from the later stages.

Agentic SAG2 raises R@10 by 8.77 points and AE@10 by 19.60 points on MuSiQue (Table~\ref{tab:agentic}). On 2WikiMultiHopQA and HotpotQA, where SAG2 is already near the ceiling, R@10 rises by 2.45 and 0.65 points. The generative model calls SAG2 2.34 times per question on MuSiQue and fewer times on 2WikiMultiHopQA and HotpotQA, where Agentic SAG2 also costs less in decision-model charges than SAG2 (Table~\ref{tab:agentic}; Appendix~\ref{app:agentic}). With the open-weight Qwen3.6-35B-A3B served locally as the agent, Agentic SAG2 lifts R@10 on MuSiQue by 5.00 points over SAG2. Under the same agent and budget, replacing SAG2 as the tool with BM25 or BGE-Large-EN-v1.5 dense retrieval (Agentic BM25 and Agentic BGE dense) lowers R@10 by 7.77 and 7.42 points. The agent's gain therefore rests on the tool it calls: each call runs the evidence scoring and term expansion of Section~\ref{sec:hypergraph}, which neither lexical nor dense retrieval supplies to the same model under the same budget.

\begin{table}[t]
\centering\footnotesize\setlength{\tabcolsep}{2.6pt}
\begin{tabularx}{\columnwidth}{@{}l>{\raggedright\arraybackslash}Xrrr@{}}
\toprule
Dataset & Method & R@10 & AE@10 & Cost (\$) \\
\midrule
MuSiQue & SAG2 & 86.62 & 69.50 & 4.33 \\
 & Agentic BM25 (Qwen3.6-35B-A3B) & 83.85$^\ddagger$ & 63.90 & -- \\
 & Agentic BGE dense (Qwen3.6-35B-A3B) & 84.20$^\ddagger$ & 65.10 & -- \\
 & Agentic SAG2 (Qwen3.6-35B-A3B) & 91.62$^\dagger$ & 79.80 & 7.80 \\
 & Agentic SAG2 (gpt-6.1-sol) & \textbf{95.39}$^\dagger$ & \textbf{89.10} & 6.85 \\
\midrule
2Wiki & SAG2 & 97.43 & 93.10 & 4.51 \\
 & Agentic SAG2 (gpt-6.1-sol) & \textbf{99.88}$^\dagger$ & \textbf{99.70} & 4.34 \\
\midrule
HotpotQA & SAG2 & 99.00 & 98.10 & 4.45 \\
 & Agentic SAG2 (gpt-6.1-sol) & \textbf{99.65}$^\dagger$ & \textbf{99.40} & 3.87 \\
\bottomrule
\end{tabularx}
\caption{Retrieval results (\%) of SAG2 and Agentic SAG2 on the 1,000-question subsets (2Wiki: 2WikiMultiHopQA), with the agent model in parentheses. Cost is the decision-model charge (\$) per 1,000 questions; -- marks tools without a decision model. R@10 differs from SAG2 on the same dataset (\(^\dagger\)) or from Agentic SAG2 (Qwen3.6-35B-A3B) (\(^\ddagger\)) at \(p<0.05\) under a paired test over questions.}\label{tab:agentic}
\end{table}

\subsection{Ablation and Analysis}\label{sec:analysis}
\begin{table}[t]
\centering\footnotesize\setlength{\tabcolsep}{3pt}
\begin{tabularx}{\columnwidth}{@{}>{\raggedright\arraybackslash}Xrrr@{}}
\toprule
Configuration & R@10 & AE@10 & Judgments \\
\midrule
BM25 & 54.50$^\dagger$ & 17.50 & 0 \\
BM25+DM, H=0 & 67.93$^\dagger$ & 36.00 & 99,925 \\
SAG2, H=1 & 83.36$^\dagger$ & 63.10 & 211,081 \\
\textbf{SAG2, H=2} & \textbf{86.62} & \textbf{69.50} & 331,325 \\
\midrule
Evidence scoring: no shared context & 76.25$^\dagger$ & 52.20 & 330,990 \\
Evidence scoring: matched BM25 budget & 69.82$^\dagger$ & 39.60 & 329,194 \\
Term activation: all seed tokens & 85.30$^\dagger$ & 66.20 & 334,036 \\
Final rescoring: new candidates only & 84.95$^\dagger$ & 66.60 & 230,684 \\
\bottomrule
\end{tabularx}
\caption{Stage contributions on the 1,000 MuSiQue questions. Upper block: the stages of Section~\ref{sec:hypergraph} added in order (BM25: lexical retrieval; BM25+DM: with evidence scoring; H: rounds of term activation and expansion, H=0 and H=1 being intermediate rankings of the H=2 run). Lower block: one stage changed with the others fixed (Section~\ref{sec:analysis}). \(^\dagger\)R@10 differs from SAG2, H=2 at \(p<0.05\) under a paired test over questions.}\label{tab:expansion}
\end{table}

\textbf{Lexical retrieval} supplies the candidate pool \(C_0\), and SAG2 needs no dense retriever. Replacing lexical retrieval with BGE dense retrieval raises the pool's R@10 by 9.55 points but the final R@10 by only 0.14 (Appendix~\ref{app:mechanism-details}), because the later stages depart from judged seeds, not from the pool's ranking (Section~\ref{sec:hypergraph}).

\textbf{Evidence scoring} judges each candidate against the question and the shared context \(E_r\). Scoring the candidate pool alone adds 13.43 R@10 points, and removing the shared context (Evidence scoring: no shared context) costs 10.37 points at an almost unchanged judgment count (Table~\ref{tab:expansion}). The context supplies the link from a candidate's fact to the chain (Section~\ref{sec:indexing}) and, through the seeds these scores select, decides which terms are expanded next. Replacing the answer-chain instruction with a topical-relevance instruction lowers R@10 by 1.75 and AE@10 by 3.00 points (Table~\ref{tab:later}): passages that name the question's entities without supplying a chain fact then become seeds, and their terms pull term expansion toward background.

\textbf{Term activation} ranks the seeds' terms by Eq.~\ref{eq:activation} and keeps 20 per round. Activating every body token of the seeds instead (Term activation: all seed tokens) lowers R@10 by 1.32 and AE@10 by 3.30 points (Table~\ref{tab:expansion}). The drop follows from the interleaving of Eq.~\ref{eq:retrieval}: the new candidates of a round are shared among all activated terms, so unranked tokens leave few slots for the terms that Eq.~\ref{eq:activation} would rank first. Multi-hop expansion runs on specific terms. Across the two expansion rounds, over three quarters of the activated terms occur in at most 58 passages (0.5\% of the corpus). Such terms account for most of the supporting passages that term expansion brings into the final top ten; their share falls from two-hop to four-hop questions as terms of medium frequency (59--200 passages) take over part of it (Table~\ref{tab:termdf} in Appendix~\ref{app:mechanism-details}). The rarity factor of Eq.~\ref{eq:weights} selects the names and phrases that a next-hop passage shares only with its seed; through them, expansion reaches passages that share no token with the question. On longer chains, the seed weight and sentence overlap of Eq.~\ref{eq:weights} rank the medium-frequency bridges that rarity alone would not place above the hundreds of other terms a seed contains.

\textbf{Term expansion} retrieves one list per activated term and interleaves the lists (Eq.~\ref{eq:retrieval}). Enlarging each question's BM25 pool to SAG2's judgment count, up to the available lexical matches, and scoring it without expansion (Evidence scoring: matched BM25 budget) leaves R@10 16.80 points below SAG2 (Table~\ref{tab:expansion}). The gain thus comes from term expansion, not from more judgments. RM3's single expanded query and GAR-BM25's precomputed neighbor graph fall behind because neither gives each bridge term its own list (Table~\ref{tab:feedback}; Section~\ref{sec:hypergraph}). Each round also changes target, from the terms of first-hop evidence to those of the second hop, whereas feedback expansion keeps aiming at the question. The second-round gain therefore grows with the hop count (Table~\ref{tab:hops}).

\textbf{Final rescoring} is a fine-grained reranking pass over the accumulated candidates with the previous top ten as context (Algorithm~\ref{alg:sag2}). Scoring only the new candidates in that round (Final rescoring: new candidates only) lowers R@10 by 1.67 and AE@10 by 2.90 points (Table~\ref{tab:expansion}). The gain comes from re-judging the earlier candidates in the completed context (Section~\ref{sec:indexing}), not from the tentative scores carried in that context: removing them changes R@10 by 0.07 points, which is not significant (Table~\ref{tab:later}).

\textbf{Scoring backend.} Five LLM decision models including Jev, four rerankers, and the embedding similarity of Qwen3-Embedding-8B \citep{zhang2025qwen3embedding} and BGE-Large-EN-v1.5 all exceed BM25 at R@10, with the LLM decision models highest (Figure~\ref{fig:dm}, Appendix~\ref{app:rerank}). A stronger decision model raises recall further: with d1-ultra \citep{liquidai2026d1ultra}, R@10 is 3.10 points above Jev. SAG2 stays above SAG with the open-weight decision models \mbox{elephant-1} (9B) and KnowLine-35B-A3B \citep{baselabs2026elephant1,pelaai2026knowline35b}. A generative LLM of the same architecture, scoring by self-reported probability or by ranked output, stays 22.28 and 9.90 points below KnowLine-35B-A3B, respectively (Appendices~\ref{app:rerank} and~\ref{app:additional}). What SAG2 requires of its scorer is therefore training to judge evidence rather than a particular model: the scores select the seeds (Section~\ref{sec:hypergraph}), so a scorer without that training misdirects term expansion as well as the final ranking.

\subsection{Efficiency and enterprise-scale retrieval}\label{sec:enterprise}
\textbf{Index and query cost.} Table~\ref{tab:efficiency} separates the one-time indexing cost from the per-question cost. The baseline rows are the runs of \citet{wu2026sag} with BGE-Large-EN-v1.5 and Qwen3.6-Flash, priced at \$0.19 and \$1.13 per million input and output tokens. The decision model of SAG2 bills only input, at \$0.042 per million, about a fifth of the LLM's input price, so indexing and retrieval on MuSiQue cost \$4.33 for SAG2 against \$14.85 for SAG. SAG2 indexes MuSiQue in seconds without calling a model, whereas the structure-augmented systems of \citet{wu2026sag} run an LLM over the corpus for hours, 3,500 to 40,000 times longer. Retrieval takes a thirteenth of SAG's time per question and an eighth of LightRAG's. SAG2 calls its decision model only at query time, so model cost grows with the number of questions rather than with the corpus (Section~\ref{sec:indexing}; batch times and usage in Appendix~\ref{app:cost}). The indexing advantage therefore widens with corpus size: the full EnterpriseRAG-Bench corpus of 511,959 documents is indexed in under eleven minutes, still without a model call (Table~\ref{tab:runs}).

\begin{table}[t]
\centering\footnotesize\setlength{\tabcolsep}{1.8pt}
\resizebox{\columnwidth}{!}{%
\begin{tabular}{@{}lrr@{\hspace{4.5pt}}rr@{\hspace{4.5pt}}r@{}}
\toprule
& \multicolumn{2}{c}{Index} & \multicolumn{2}{c}{Retrieval} & \\
\cmidrule(lr){2-3}\cmidrule(lr){4-5}
Method & Time & Tokens (M) & Time (s/q) & Tokens (M) & Cost (\$) \\
\midrule
RAG (BGE) & 0.00\,h & 0 & 0.10 & 0 & 0.00 \\
GraphRAG & 10.15\,h & 125.12 & 0.23 & 0 & 50.55 \\
LightRAG & 25.94\,h & 103.14 & 11.97 & 0.52 & 41.77 \\
HyperGraphRAG & 16.21\,h & 216.11 & 4.45 & 0.50 & 61.46 \\
HippoRAG 2 & 2.30\,h & 15.93 & 0.66 & 3.05 & 7.94 \\
SAG & 4.57\,h & 28.26 & 18.67 & 10.17 & 14.85 \\
\midrule
SAG2 & 2.36\,s & 0 & 1.47 & 109.98\,(DM) & 4.33 \\
\bottomrule
\end{tabular}}
\caption{Indexing and retrieval cost on MuSiQue. Tokens are model input plus output in millions; retrieval tokens are totals over the 1,000 questions, and Cost covers indexing plus their retrieval. Baseline rows are as reported by \citet{wu2026sag}; DM marks decision-model tokens. No row includes answer generation.}\label{tab:efficiency}
\end{table}

\textbf{Enterprise documents.} On the full corpus of EnterpriseRAG-Bench, SAG2 reaches 81.12\% document R@10, above the BM25 and vector retrieval results reported with the benchmark \citep{sun2026enterpriserag} (Table~\ref{tab:enterprise} in Appendix~\ref{app:scaling}). The benchmark's own BM25 retriever is the stronger of its two baselines on this corpus. SAG2 surpasses it from the same kind of lexical index, so the margin comes from the decision-model stages rather than from a different retriever.

\begin{figure}[!t]
\centering
\includegraphics[width=\columnwidth]{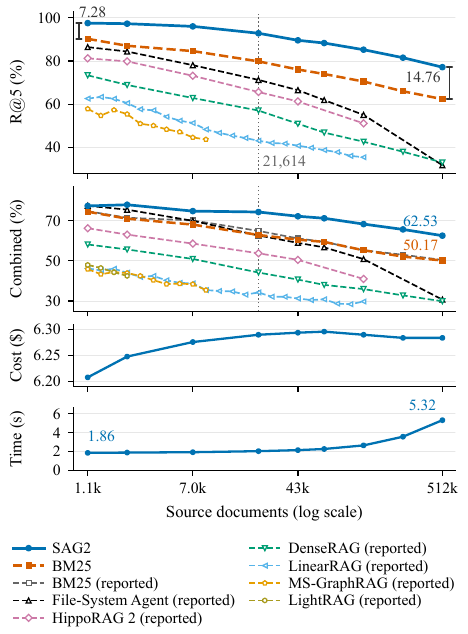}
\caption{Corpus growth over nine nested tiers of EnterpriseRAG-Bench, with recall measured on the 470 labeled questions and cost and time on all 500 questions. First panel: document R@5 of SAG2 and BM25, with the document recall reported by \citet{wang2026bm25scale} for five other systems at the tiers they report; brackets mark the SAG2--BM25 gap at the first and last tiers, and the dotted line 21,614 documents. Second panel: official combined score, with SAG2 and BM25 as our runs under the same reader, judge, and reader input length and the other curves as reported by \citet{wang2026bm25scale}. Third and fourth panels: decision-model cost per 1,000 questions and mean end-to-end time per question; the cost axis starts above zero.}\label{fig:scaling}
\end{figure}

\textbf{Corpus growth.} As a corpus grows, more background documents share query terms with the evidence. Following the nested-corpus design of \citet{wang2026bm25scale}, we index EnterpriseRAG-Bench at the nine tiers at which they report BM25, from 1,144 to 511,959 source documents. Every tier contains all reference documents of the 470 questions and is indexed separately (Appendix~\ref{app:scaling}). Over this 447.5-fold growth, SAG2's R@5 lead over BM25 widens from 7.28 to 14.76 points (Figure~\ref{fig:scaling}). The widening lead therefore comes from the evidence scoring and term expansion that SAG2 adds to lexical matching. It grows as background documents accumulate because documents that share query terms disturb judgment-guided expansion less than they disturb lexical matching alone. Beyond 21,614 documents, the gap stays between 13.00 and 15.54 points (Appendix~\ref{app:scaling}). At every tier, SAG2 is also above every other method that \citet{wang2026bm25scale} report. Decision-model cost per 1,000 questions varies by 1.4 percent across the tiers, since the retrieval budget, not the corpus size, bounds the judgments. Mean end-to-end time per question rises about 2.9-fold over the same growth, as lexical retrieval searches a larger index. The model cost of SAG2 can therefore be budgeted from its parameters before the corpus is known, and its time per question grows slowly with the corpus. Answer quality under the same growth, with the reader and judge of \citet{wang2026bm25scale} and as much reader input as BM25 supplies, stays above BM25 at every tier (Figure~\ref{fig:scaling}, second panel; Appendix~\ref{app:scaling}).

\FloatBarrier
\section{Conclusion}
Multi-hop retrieval must reach passages that become useful only after intermediate evidence is retrieved. SAG2 reaches them by letting the evidence judged so far decide which lexical connections to follow: a decision model scores each candidate's contribution to the answer chain in a shared context, and the terms activated from the highest-scoring seeds retrieve unvisited passages through the lexical index. That index, built without embeddings or an LLM, is the only persistent structure. It holds a passage--term hypergraph implicitly, and retrieval computes only the local hypergraph that a question's judged seeds reach. With this design, SAG2 reaches the highest R@10 among the compared systems on MuSiQue, 2WikiMultiHopQA, and HotpotQA. It also leads on EnterpriseRAG-Bench, where its margin over BM25 alone widens over 447.5-fold corpus growth at a nearly constant model cost per question. Deployment needs only a SQLite index and a decision model, so SAG2 suits corpora that change often or must be searchable as soon as they are indexed.

\section*{Limitations}
Indexing without an LLM moves model computation to query time, where each question pays for its own decision-model judgments, and on a stable corpus these charges accumulate. With indexing and retrieval counted at list prices, without answer generation, and with the baseline usage of \citet[Table~8]{wu2026sag}, SAG2's cumulative cost on MuSiQue exceeds that of HippoRAG~2 after about 1,976 questions and that of SAG after about 10,216 (Table~\ref{tab:efficiency}).

Retrieval quality depends on the decision model (Section~\ref{sec:analysis}), and the default, Jev, is a hosted service whose weights are not released. The open-weight decision models of Section~\ref{sec:analysis}, elephant-1 and KnowLine-35B-A3B, replace it while keeping SAG2 above SAG. Deploying and serving them, however, currently takes more hardware and engineering than calling the hosted model, because these models have not yet been optimized for inference efficiency. This overhead falls as open-weight decision models are optimized.

Of MuSiQue's supporting passages, 1.9\% share no content word with the question or any activated term, and others fall outside the list depth or the interleaved merge (Table~\ref{tab:loss} in Appendix~\ref{app:supp}). On the four-hop questions whose last-hop passage never enters a candidate pool, 88.1\% of these passages contain none of the activated terms. The activation budget, not lexical ranking, thus bounds these chains, and a larger budget reaches the bridge term at the cost of more judgments. Injected into the final pool with the context unchanged, 54.8\% of these last-hop passages enter the top ten (Appendix~\ref{app:supp}), so retrieving them would bring more than half into the final ranking.

The share of supporting passages rescored but ranked below ten rises to 11.1\% on four-hop questions, where the margin below the tenth widens. The 45.2\% of injected last-hop passages that stay below the tenth candidate point to the same limit: on these four-hop chains, the decision model misses a last link that connects to the question only through the context. On four-hop questions, about a quarter of the lost passages follow an earlier lost link of the chain.

\bibliography{references}

\begin{thebibliography}{89}
\providecommand{\natexlab}[1]{#1}

\bibitem[{Abdul-Jaleel et~al.(2004)Abdul-Jaleel, Allan, Croft, Diaz, Larkey,
  Li, Smucker, and Wade}]{abduljaleel2004rm3}
Nasreen Abdul-Jaleel, James Allan, W.~Bruce Croft, Fernando Diaz, Leah~S.
  Larkey, Xiaoyan Li, Mark~D. Smucker, and Courtney Wade. 2004.
\newblock \href {https://doi.org/10.6028/nist.sp.500-261.hard-u.mass} {{UMass}
  at {TREC} 2004: Novelty and {HARD}}.
\newblock In \emph{Proceedings of the Thirteenth Text {RE}trieval Conference
  ({TREC} 2004)}, NIST Special Publication 500-261. National Institute of
  Standards and Technology (NIST).

\bibitem[{Almeida(2026)}]{almeida2026jev}
Diogo Almeida. 2026.
\newblock \href
  {https://typesafe.ai/blog/introducing-system-one-models-and-jev} {Introducing
  {System One} models \& {Jev}}.
\newblock TypeSafe AI Blog.
\newblock Published 15 September 2026. Accessed 2026-10-19.

\bibitem[{An et~al.(2026)An, Quang, An, Van, Le, and Nguyen}]{an2026hyperprove}
Nguyen~Phu An, Dung~Nguyen Quang, Luu~Hieu An, Linh~Ngo Van, Trung Le, and
  Thien~Huu Nguyen. 2026.
\newblock \href {https://doi.org/10.18653/2026.emnlp-main.212} {{H}yper{P}rove:
  Answer-guided hypergraph expansion for multi-hop question answering}.
\newblock In \emph{Proceedings of the 2026 Conference on Empirical Methods in
  Natural Language Processing}, pages 4867--4882, Budapest, Hungary.
  Association for Computational Linguistics.

\bibitem[{Bacellar(2026)}]{bacellar2026bridgerag}
Andre Bacellar. 2026.
\newblock \href {https://arxiv.org/abs/2604.03384} {{BridgeRAG}: Training-free
  bridge-conditioned retrieval for multi-hop question answering}.
\newblock \emph{Preprint}, arXiv:2604.03384.

\bibitem[{{Base Labs}(2026)}]{baselabs2026elephant1}
{Base Labs}. 2026.
\newblock \href {https://huggingface.co/baselabs/elephant-1} {elephant-1 model
  card}.
\newblock Hugging Face.
\newblock 9B decision model fine-tuned from Qwen3.5-9B; model weights under the
  MIT license. Accessed 2026-11-03.

\bibitem[{Bigdeli et~al.(2026)Bigdeli, Hamidi~Rad, Arabzadeh, Ebrahimi, Le,
  Clarke, and Bagheri}]{bigdeli2026seek}
Amin Bigdeli, Radin Hamidi~Rad, Negar Arabzadeh, Sajad Ebrahimi, Hai~Son Le,
  Charles L.~A. Clarke, and Ebrahim Bagheri. 2026.
\newblock \href {https://arxiv.org/abs/2609.28980} {Seek: Self-evaluative
  exploration for knowledge retrieval}.
\newblock \emph{Preprint}, arXiv:2609.28980.
\newblock Accepted at CIKM 2026.

\bibitem[{Bonifazi et~al.(2026)Bonifazi, Buratti, Marchetti, Parlapiano,
  Quaglieri, Traini, Ursino, and Virgili}]{bonifazi2026tigrag}
Gianluca Bonifazi, Christopher Buratti, Michele Marchetti, Federica Parlapiano,
  Giulia Quaglieri, Davide Traini, Domenico Ursino, and Luca Virgili. 2026.
\newblock \href {https://arxiv.org/abs/2606.30093} {Efficient
  retrieval-augmented generation via token co-occurrence graphs}.
\newblock \emph{Preprint}, arXiv:2606.30093.

\bibitem[{Champclaux et~al.(2009)Champclaux, Dkaki, and
  Mothe}]{champclaux2009okasim}
Ya{\"e}l Champclaux, Taoufiq Dkaki, and Josiane Mothe. 2009.
\newblock \href {https://doi.org/10.5220/0002017202790285} {Enhancing high
  precision by combining {Okapi} {BM25} with structural similarity in an
  information retrieval system}.
\newblock In \emph{Proceedings of the 11th International Conference on
  Enterprise Information Systems ({ICEIS}) -- Information Systems Analysis and
  Specification}, pages 279--285. SciTePress.

\bibitem[{Cormack et~al.(2009)Cormack, Clarke, and Buettcher}]{cormack2009rrf}
Gordon~V. Cormack, Charles L.~A. Clarke, and Stefan Buettcher. 2009.
\newblock \href {https://doi.org/10.1145/1571941.1572114} {Reciprocal rank
  fusion outperforms {Condorcet} and individual rank learning methods}.
\newblock In \emph{Proceedings of the 32nd International {ACM} {SIGIR}
  Conference on Research and Development in Information Retrieval}, SIGIR '09,
  pages 758--759. ACM.

\bibitem[{Das et~al.(2019)Das, Godbole, Kavarthapu, Gong, Singhal, Yu, Guo,
  Gao, Zamani, Zaheer, and McCallum}]{das2019multistep}
Rajarshi Das, Ameya Godbole, Dilip Kavarthapu, Zhiyu Gong, Abhishek Singhal,
  Mo~Yu, Xiaoxiao Guo, Tian Gao, Hamed Zamani, Manzil Zaheer, and Andrew
  McCallum. 2019.
\newblock \href {https://doi.org/10.18653/v1/D19-5816} {Multi-step
  entity-centric information retrieval for multi-hop question answering}.
\newblock In \emph{Proceedings of the 2nd Workshop on Machine Reading for
  Question Answering}, pages 113--118, Hong Kong, China. Association for
  Computational Linguistics.

\bibitem[{Devezas and Nunes(2019)}]{devezas2019hypergraphofentity}
Jos{\'e} Devezas and S{\'e}rgio Nunes. 2019.
\newblock \href {https://doi.org/10.1515/comp-2019-0006} {Hypergraph-of-entity:
  A unified representation model for the retrieval of text and knowledge}.
\newblock \emph{Open Computer Science}, 9(1):103--127.

\bibitem[{Ding et~al.(2019)Ding, Zhou, Chen, Yang, and Tang}]{ding2019cogqa}
Ming Ding, Chang Zhou, Qibin Chen, Hongxia Yang, and Jie Tang. 2019.
\newblock \href {https://doi.org/10.18653/v1/P19-1259} {Cognitive graph for
  multi-hop reading comprehension at scale}.
\newblock In \emph{Proceedings of the 57th Annual Meeting of the {A}ssociation
  for {C}omputational {L}inguistics}, pages 2694--2703, Florence, Italy.
  Association for Computational Linguistics.

\bibitem[{Edge et~al.(2024)Edge, Trinh, and Larson}]{edge2024lazygraphrag}
Darren Edge, Ha~Trinh, and Jonathan Larson. 2024.
\newblock \href
  {https://www.microsoft.com/en-us/research/blog/lazygraphrag-setting-a-new-standard-for-quality-and-cost/}
  {{LazyGraphRAG}: Setting a new standard for quality and cost}.
\newblock Microsoft Research Blog, November 25, 2024.

\bibitem[{Efron and Tibshirani(1993)}]{efron1993bootstrap}
Bradley Efron and Robert~J. Tibshirani. 1993.
\newblock \href {https://doi.org/10.1007/978-1-4899-4541-9} {\emph{An
  Introduction to the Bootstrap}}.
\newblock Springer US, Boston, MA.

\bibitem[{Feng et~al.(2026)Feng, Hu, Ying, Hou, Liu, Yang, Li, Du, Zheng, Hu,
  and Gao}]{feng2026hyperrag}
Yifan Feng, Hao Hu, Shihui Ying, Xingliang Hou, Shiquan Liu, Mingyuan Yang,
  Junchang Li, Shaoyi Du, Nanning Zheng, Han Hu, and Yue Gao. 2026.
\newblock \href {https://doi.org/10.1038/s41467-026-71411-1} {{Hyper-RAG}:
  Combating {LLM} hallucinations using hypergraph-driven retrieval-augmented
  generation}.
\newblock \emph{Nature Communications}, 17:5778.

\bibitem[{Fernandes and Pereira(2026)}]{fernandes2026jevjudge}
Leandro~Car{\'i}sio Fernandes and Jayr Pereira. 2026.
\newblock \href {https://arxiv.org/abs/2610.23346} {Jev-as-a-judge for
  {Brazilian} {Portuguese} information retrieval: Relevance assessment and
  reranking}.
\newblock \emph{Preprint}, arXiv:2610.23346.

\bibitem[{Gao et~al.(2023)Gao, Ma, Lin, and Callan}]{gao2023hyde}
Luyu Gao, Xueguang Ma, Jimmy Lin, and Jamie Callan. 2023.
\newblock \href {https://doi.org/10.18653/v1/2023.acl-long.99} {Precise
  zero-shot dense retrieval without relevance labels}.
\newblock In \emph{Proceedings of the 61st Annual Meeting of the {A}ssociation
  for {C}omputational {L}inguistics (Volume 1: Long Papers)}, pages 1762--1777,
  Toronto, Canada. Association for Computational Linguistics.

\bibitem[{Guti\'{e}rrez et~al.(2024)Guti\'{e}rrez, Shu, Gu, Yasunaga, and
  Su}]{gutierrez2024hipporag}
Bernal~Jim\'{e}nez Guti\'{e}rrez, Yiheng Shu, Yu~Gu, Michihiro Yasunaga, and
  Yu~Su. 2024.
\newblock \href {https://doi.org/10.52202/079017-1902} {{HippoRAG}:
  Neurobiologically inspired long-term memory for large language models}.
\newblock In \emph{Advances in Neural Information Processing Systems},
  volume~37, pages 59532--59569. Curran Associates, Inc.

\bibitem[{Guti\'{e}rrez et~al.(2025)Guti\'{e}rrez, Shu, Qi, Zhou, and
  Su}]{gutierrez2025hipporag2}
Bernal~Jim\'{e}nez Guti\'{e}rrez, Yiheng Shu, Weijian Qi, Sizhe Zhou, and
  Yu~Su. 2025.
\newblock \href {https://proceedings.mlr.press/v267/gutierrez25a.html} {From
  {RAG} to memory: Non-parametric continual learning for large language
  models}.
\newblock In \emph{Proceedings of the 42nd International Conference on Machine
  Learning}, volume 267 of \emph{Proceedings of Machine Learning Research},
  pages 21497--21515. PMLR.

\bibitem[{Ho et~al.(2020)Ho, Duong~Nguyen, Sugawara, and
  Aizawa}]{ho2020twowiki}
Xanh Ho, Anh-Khoa Duong~Nguyen, Saku Sugawara, and Akiko Aizawa. 2020.
\newblock \href {https://doi.org/10.18653/v1/2020.coling-main.580}
  {Constructing a multi-hop {QA} dataset for comprehensive evaluation of
  reasoning steps}.
\newblock In \emph{Proceedings of the 28th International Conference on
  Computational Linguistics}, pages 6609--6625, Barcelona, Spain (Online).
  International Committee on Computational Linguistics.

\bibitem[{Huang et~al.(2025)Huang, Zhang, and Xiao}]{huang2025ketrag}
Yiqian Huang, Shiqi Zhang, and Xiaokui Xiao. 2025.
\newblock \href {https://doi.org/10.1145/3711896.3737012} {{KET-RAG}: A
  cost-efficient multi-granular indexing framework for {Graph-RAG}}.
\newblock In \emph{Proceedings of the 31st {ACM} {SIGKDD} Conference on
  Knowledge Discovery and Data Mining V.2}, KDD '25, pages 1003--1012. ACM.

\bibitem[{Izacard et~al.(2022)Izacard, Caron, Hosseini, Riedel, Bojanowski,
  Joulin, and Grave}]{izacard2022contriever}
Gautier Izacard, Mathilde Caron, Lucas Hosseini, Sebastian Riedel, Piotr
  Bojanowski, Armand Joulin, and Edouard Grave. 2022.
\newblock \href {https://openreview.net/forum?id=jKN1pXi7b0} {Unsupervised
  dense information retrieval with contrastive learning}.
\newblock \emph{Transactions on Machine Learning Research}.

\bibitem[{Jin et~al.(2025)Jin, Zeng, Yue, Yoon, Arik, Wang, Zamani, and
  Han}]{jin2025searchr1}
Bowen Jin, Hansi Zeng, Zhenrui Yue, Jinsung Yoon, Sercan Arik, Dong Wang, Hamed
  Zamani, and Jiawei Han. 2025.
\newblock \href {https://openreview.net/forum?id=Rwhi91ideu} {{Search-R1}:
  Training {LLM}s to reason and leverage search engines with reinforcement
  learning}.
\newblock In \emph{Second Conference on Language Modeling}.

\bibitem[{{Jina AI}(2026)}]{jinaai2026rerankerv35}
{Jina AI}. 2026.
\newblock \href {https://huggingface.co/jinaai/jina-reranker-v3.5}
  {jina-reranker-v3.5 model card}.
\newblock Hugging Face.
\newblock Model weights, local inference and CC BY-NC 4.0 license. Accessed
  2026-10-16.

\bibitem[{Karpukhin et~al.(2020)Karpukhin, Oguz, Min, Lewis, Wu, Edunov, Chen,
  and Yih}]{karpukhin2020dpr}
Vladimir Karpukhin, Barlas Oguz, Sewon Min, Patrick Lewis, Ledell Wu, Sergey
  Edunov, Danqi Chen, and Wen-tau Yih. 2020.
\newblock \href {https://doi.org/10.18653/v1/2020.emnlp-main.550} {Dense
  passage retrieval for open-domain question answering}.
\newblock In \emph{Proceedings of the 2020 Conference on Empirical Methods in
  Natural Language Processing ({EMNLP})}, pages 6769--6781, Online. Association
  for Computational Linguistics.

\bibitem[{Khattab et~al.(2021)Khattab, Potts, and Zaharia}]{khattab2021baleen}
Omar Khattab, Christopher Potts, and Matei Zaharia. 2021.
\newblock \href
  {https://proceedings.neurips.cc/paper/2021/hash/e8b1cbd05f6e6a358a81dee52493dd06-Abstract.html}
  {Baleen: Robust multi-hop reasoning at scale via condensed retrieval}.
\newblock In \emph{Advances in Neural Information Processing Systems},
  volume~34, pages 27670--27682. Curran Associates, Inc.

\bibitem[{Kim et~al.(2021)Kim, Jeong, Choi, and
  Hwang}]{kim2021structurekeyphrase}
Jihyuk Kim, Myeongho Jeong, Seungtaek Choi, and Seung-won Hwang. 2021.
\newblock \href {https://doi.org/10.18653/v1/2021.emnlp-main.209}
  {Structure-augmented keyphrase generation}.
\newblock In \emph{Proceedings of the 2021 Conference on Empirical Methods in
  Natural Language Processing}, pages 2657--2667, Online and Punta Cana,
  Dominican Republic. Association for Computational Linguistics.

\bibitem[{Kim et~al.(2026)Kim, Kim, Kim, Kim, Hwang, and Lee}]{acl2026repair}
Jongho Kim, Jaeyoung Kim, Jihyuk Kim, Yu~Jin Kim, Seung-won Hwang, and Moontae
  Lee. 2026.
\newblock \href {https://doi.org/10.18653/v1/2026.acl-long.1734} {Adaptive
  retrieval for reasoning}.
\newblock In \emph{Proceedings of the 64th Annual Meeting of the {A}ssociation
  for {C}omputational {L}inguistics (Volume 1: Long Papers)}, pages
  37366--37381, San Diego, California, United States. Association for
  Computational Linguistics.
\newblock ArXiv:2601.04618.

\bibitem[{Koehn(2004)}]{koehn2004significance}
Philipp Koehn. 2004.
\newblock \href {https://aclanthology.org/W04-3250/} {Statistical significance
  tests for machine translation evaluation}.
\newblock In \emph{Proceedings of the 2004 Conference on Empirical Methods in
  Natural Language Processing}, pages 388--395, Barcelona, Spain. Association
  for Computational Linguistics.

\bibitem[{Kurland and Lee(2005)}]{kurland2005pagerank}
Oren Kurland and Lillian Lee. 2005.
\newblock \href {https://doi.org/10.1145/1076034.1076087} {{PageRank} without
  hyperlinks: Structural re-ranking using links induced by language models}.
\newblock In \emph{Proceedings of the 28th Annual International {ACM} {SIGIR}
  Conference on Research and Development in Information Retrieval}, SIGIR '05,
  pages 306--313. ACM.

\bibitem[{Lavrenko and Croft(2001)}]{lavrenko2001relevance}
Victor Lavrenko and W.~Bruce Croft. 2001.
\newblock \href {https://doi.org/10.1145/383952.383972} {Relevance based
  language models}.
\newblock In \emph{Proceedings of the 24th Annual International {ACM} {SIGIR}
  Conference on Research and Development in Information Retrieval}, SIGIR '01,
  pages 120--127. ACM.

\bibitem[{Lee et~al.(2025)Lee, Roy, Xu, Raiman, Shoeybi, Catanzaro, and
  Ping}]{lee2025nvembed}
Chankyu Lee, Rajarshi Roy, Mengyao Xu, Jonathan Raiman, Mohammad Shoeybi, Bryan
  Catanzaro, and Wei Ping. 2025.
\newblock \href
  {https://proceedings.iclr.cc/paper_files/paper/2025/hash/c4bf73386022473a652a18941e9ea6f8-Abstract-Conference.html}
  {{NV-Embed}: Improved techniques for training {LLM}s as generalist embedding
  models}.
\newblock In \emph{International Conference on Learning Representations}, pages
  79310--79333.

\bibitem[{Lei et~al.(2024)Lei, Cao, Zhou, Shen, and Yates}]{lei2024csqe}
Yibin Lei, Yu~Cao, Tianyi Zhou, Tao Shen, and Andrew Yates. 2024.
\newblock \href {https://doi.org/10.18653/v1/2024.eacl-short.34}
  {Corpus-steered query expansion with large language models}.
\newblock In \emph{Proceedings of the 18th Conference of the {E}uropean Chapter
  of the {A}ssociation for {C}omputational {L}inguistics (Volume 2: Short
  Papers)}, pages 393--401, St. Julian's, Malta. Association for Computational
  Linguistics.

\bibitem[{Li et~al.(2025)Li, Dong, Jin, Zhang, Zhou, Zhu, Zhang, and
  Dou}]{li2025searcho1}
Xiaoxi Li, Guanting Dong, Jiajie Jin, Yuyao Zhang, Yujia Zhou, Yutao Zhu,
  Peitian Zhang, and Zhicheng Dou. 2025.
\newblock \href {https://doi.org/10.18653/v1/2025.emnlp-main.276} {{S}earch-o1:
  Agentic search-enhanced large reasoning models}.
\newblock In \emph{Proceedings of the 2025 Conference on Empirical Methods in
  Natural Language Processing}, pages 5420--5438, Suzhou, China. Association
  for Computational Linguistics.

\bibitem[{Li et~al.(2023)Li, Zhang, Zhang, Long, Xie, and Zhang}]{li2023gte}
Zehan Li, Xin Zhang, Yanzhao Zhang, Dingkun Long, Pengjun Xie, and Meishan
  Zhang. 2023.
\newblock \href {https://arxiv.org/abs/2308.03281} {Towards general text
  embeddings with multi-stage contrastive learning}.
\newblock \emph{Preprint}, arXiv:2308.03281.

\bibitem[{Li et~al.(2026)Li, Zhang, Wei, Lu, Nie, Lu, Bai, Feng, Zhu, Zhong,
  Zhang, Xie, Choi, Zou, Han, Chen, Lin, Jiang, and Zhang}]{li2026dci}
Zhuofeng Li, Haoxiang Zhang, Cong Wei, Pan Lu, Ping Nie, Yi~Lu, Yuyang Bai,
  Shangbin Feng, Hangxiao Zhu, Ming Zhong, Yuyu Zhang, Jianwen Xie, Yejin Choi,
  James Zou, Jiawei Han, Wenhu Chen, Jimmy Lin, Dongfu Jiang, and Yu~Zhang.
  2026.
\newblock \href {https://arxiv.org/abs/2605.05242} {Beyond semantic similarity:
  Rethinking retrieval for agentic search via direct corpus interaction}.
\newblock \emph{Preprint}, arXiv:2605.05242.

\bibitem[{{Liquid AI}(2026)}]{liquidai2026d1ultra}
{Liquid AI}. 2026.
\newblock \href {https://www.liquid.ai/blog/d1-ultra} {d1-ultra: Introducing
  the world's best decision model}.
\newblock Liquid AI blog.
\newblock Hosted decision model released 2026-10-15; weights not published.
  Accessed 2026-11-11.

\bibitem[{Liu et~al.(2025)Liu, Wang, Chen, Li, Xiong, Yu, and
  Zhang}]{liu2025hoprag}
Hao Liu, Zhengren Wang, Xi~Chen, Zhiyu Li, Feiyu Xiong, Qinhan Yu, and Wentao
  Zhang. 2025.
\newblock \href {https://doi.org/10.18653/v1/2025.findings-acl.97} {{H}op{RAG}:
  Multi-hop reasoning for logic-aware retrieval-augmented generation}.
\newblock In \emph{Findings of the {A}ssociation for {C}omputational
  {L}inguistics: {ACL} 2025}, pages 1897--1913, Vienna, Austria. Association
  for Computational Linguistics.

\bibitem[{Liu et~al.(2026)Liu, Chen, Yang, He, Ju, Han, Liu, and
  Zhou}]{liu2026hypersu}
Jiate Liu, Liuyi Chen, Zhengyi Yang, Chuan He, Mingchen Ju, Bocheng Han, Ruyi
  Liu, and Xu~Zhou. 2026.
\newblock \href {https://doi.org/10.18653/2026.findings-emnlp.1058}
  {{H}yper{SU}: Corpus-driven semantic-unit hypergraph for retrieval-augmented
  generation}.
\newblock In \emph{Findings of the {A}ssociation for {C}omputational
  {L}inguistics: {EMNLP} 2026}, pages 22399--22424, Budapest, Hungary.
  Association for Computational Linguistics.

\bibitem[{Luo et~al.(2025)Luo, E, Chen, Zheng, Wu, Guo, Lin, Feng, Kuang, Song,
  Zhu, and Luu}]{luo2025hypergraphrag}
Haoran Luo, Haihong E, Guanting Chen, Yandan Zheng, Xiaobao Wu, Yikai Guo, Qika
  Lin, Yu~Feng, Zemin Kuang, Meina Song, Yifan Zhu, and Anh~Tuan Luu. 2025.
\newblock \href {https://doi.org/10.52202/085713-5089} {{HyperGraphRAG}:
  Retrieval-augmented generation via hypergraph-structured knowledge
  representation}.
\newblock In \emph{Advances in Neural Information Processing Systems},
  volume~38, pages 152206--152234. Curran Associates, Inc.

\bibitem[{MacAvaney et~al.(2022)MacAvaney, Tonellotto, and
  Macdonald}]{macavaney2022gar}
Sean MacAvaney, Nicola Tonellotto, and Craig Macdonald. 2022.
\newblock \href {https://doi.org/10.1145/3511808.3557231} {Adaptive re-ranking
  with a corpus graph}.
\newblock In \emph{Proceedings of the 31st {ACM} International Conference on
  Information \& Knowledge Management}, CIKM '22, pages 1491--1500. ACM.

\bibitem[{Muennighoff et~al.(2025)Muennighoff, Su, Wang, Yang, Wei, Yu, Singh,
  and Kiela}]{muennighoff2024gritlm}
Niklas Muennighoff, Hongjin Su, Liang Wang, Nan Yang, Furu Wei, Tao Yu,
  Amanpreet Singh, and Douwe Kiela. 2025.
\newblock \href
  {https://proceedings.iclr.cc/paper_files/paper/2025/hash/70cf215430492f7d34830a24e744b3f1-Abstract-Conference.html}
  {Generative representational instruction tuning}.
\newblock In \emph{International Conference on Learning Representations}, pages
  45544--45613.

\bibitem[{Nogueira and Cho(2019)}]{nogueira2019passage}
Rodrigo Nogueira and Kyunghyun Cho. 2019.
\newblock \href {https://arxiv.org/abs/1901.04085} {Passage re-ranking with
  {BERT}}.
\newblock \emph{Preprint}, arXiv:1901.04085.

\bibitem[{Nogueira et~al.(2020)Nogueira, Jiang, Pradeep, and
  Lin}]{nogueira2020monot5}
Rodrigo Nogueira, Zhiying Jiang, Ronak Pradeep, and Jimmy Lin. 2020.
\newblock \href {https://doi.org/10.18653/v1/2020.findings-emnlp.63} {Document
  ranking with a pretrained sequence-to-sequence model}.
\newblock In \emph{Findings of the {A}ssociation for {C}omputational
  {L}inguistics: {EMNLP} 2020}, pages 708--718, Online. Association for
  Computational Linguistics.

\bibitem[{Otero and Parapar(2026)}]{otero2026llmprf}
David Otero and Javier Parapar. 2026.
\newblock \href {https://doi.org/10.1007/978-3-032-21300-6_36} {{LLM}-assisted
  pseudo-relevance feedback}.
\newblock In \emph{Advances in Information Retrieval ({ECIR} 2026)}, volume
  16484 of \emph{Lecture Notes in Computer Science}, pages 452--459, Cham.
  Springer Nature Switzerland.

\bibitem[{Park et~al.(2026)Park, Kwon, Kim, and Lee}]{park2026garen}
June Park, Jun Kwon, Jonghyo Kim, and Jongwuk Lee. 2026.
\newblock \href {https://doi.org/10.18653/2026.emnlp-main.2143} {Group-aware
  adaptive retrieval for evidence navigation}.
\newblock In \emph{Proceedings of the 2026 Conference on Empirical Methods in
  Natural Language Processing}, pages 48728--48747, Budapest, Hungary.
  Association for Computational Linguistics.

\bibitem[{{PelaAI}(2026{\natexlab{a}})}]{pelaai2026knowline35b}
{PelaAI}. 2026{\natexlab{a}}.
\newblock \href {https://huggingface.co/PelaAI/KnowLine-35B-A3B-Gen1.5}
  {{KnowLine-35B-A3B-Gen1.5} model card}.
\newblock Hugging Face.
\newblock Decision model fine-tuned from Qwen3.5-35B-A3B; model weights under
  the Apache-2.0 license. Accessed 2026-11-03.

\bibitem[{{PelaAI}(2026{\natexlab{b}})}]{pelaai2026knowline4b}
{PelaAI}. 2026{\natexlab{b}}.
\newblock \href {https://huggingface.co/PelaAI/KnowLine-4B-Gen5}
  {{KnowLine-4B-Gen5} model card}.
\newblock Hugging Face.
\newblock Decision model fine-tuned from Qwen3.5-4B; model weights under the
  Apache-2.0 license. Accessed 2026-11-03.

\bibitem[{Press et~al.(2023)Press, Zhang, Min, Schmidt, Smith, and
  Lewis}]{press2023selfask}
Ofir Press, Muru Zhang, Sewon Min, Ludwig Schmidt, Noah Smith, and Mike Lewis.
  2023.
\newblock \href {https://doi.org/10.18653/v1/2023.findings-emnlp.378}
  {Measuring and narrowing the compositionality gap in language models}.
\newblock In \emph{Findings of the {A}ssociation for {C}omputational
  {L}inguistics: {EMNLP} 2023}, pages 5687--5711, Singapore. Association for
  Computational Linguistics.

\bibitem[{Qi et~al.(2019)Qi, Lin, Mehr, Wang, and Manning}]{qi2019golden}
Peng Qi, Xiaowen Lin, Leo Mehr, Zijian Wang, and Christopher~D. Manning. 2019.
\newblock \href {https://doi.org/10.18653/v1/D19-1261} {Answering complex
  open-domain questions through iterative query generation}.
\newblock In \emph{Proceedings of the 2019 Conference on Empirical Methods in
  Natural Language Processing and the 9th International Joint Conference on
  Natural Language Processing ({EMNLP}-{IJCNLP})}, pages 2590--2602, Hong Kong,
  China. Association for Computational Linguistics.

\bibitem[{{Qwen Team}(2025)}]{qwen2025reranker06b}
{Qwen Team}. 2025.
\newblock \href {https://huggingface.co/Qwen/Qwen3-Reranker-0.6B}
  {{Qwen3-Reranker-0.6B} model card}.
\newblock Hugging Face.
\newblock Model weights, local inference and Apache-2.0 license. Accessed
  2026-10-16.

\bibitem[{Rathee et~al.(2025)Rathee, MacAvaney, and Anand}]{rathee2025slidegar}
Mandeep Rathee, Sean MacAvaney, and Avishek Anand. 2025.
\newblock \href {https://doi.org/10.1007/978-3-031-88708-6_15} {Guiding
  retrieval using {LLM}-based listwise rankers}.
\newblock In \emph{Advances in Information Retrieval ({ECIR} 2025)}, volume
  15572 of \emph{Lecture Notes in Computer Science}, pages 230--246, Cham.
  Springer Nature Switzerland.

\bibitem[{Robertson and Zaragoza(2009)}]{robertson2009bm25}
Stephen Robertson and Hugo Zaragoza. 2009.
\newblock \href {https://doi.org/10.1561/1500000019} {The probabilistic
  relevance framework: {BM25} and beyond}.
\newblock \emph{Foundations and Trends in Information Retrieval},
  3(4):333--389.

\bibitem[{Rousseau and Vazirgiannis(2013)}]{rousseau2013graphofword}
Fran{\c{c}}ois Rousseau and Michalis Vazirgiannis. 2013.
\newblock \href {https://doi.org/10.1145/2505515.2505671} {Graph-of-word and
  {TW-IDF}: New approach to ad hoc {IR}}.
\newblock In \emph{Proceedings of the 22nd {ACM} International Conference on
  Information and Knowledge Management}, CIKM '13, pages 59--68. ACM.

\bibitem[{Shao et~al.(2023)Shao, Gong, Shen, Huang, Duan, and
  Chen}]{shao2023iterretgen}
Zhihong Shao, Yeyun Gong, Yelong Shen, Minlie Huang, Nan Duan, and Weizhu Chen.
  2023.
\newblock \href {https://doi.org/10.18653/v1/2023.findings-emnlp.620}
  {Enhancing retrieval-augmented large language models with iterative
  retrieval-generation synergy}.
\newblock In \emph{Findings of the {A}ssociation for {C}omputational
  {L}inguistics: {EMNLP} 2023}, pages 9248--9274, Singapore. Association for
  Computational Linguistics.

\bibitem[{Shen et~al.(2024)Shen, Long, Geng, Tao, Lei, Zhou, Blumenstein, and
  Jiang}]{shen2024lamer}
Tao Shen, Guodong Long, Xiubo Geng, Chongyang Tao, Yibin Lei, Tianyi Zhou,
  Michael Blumenstein, and Daxin Jiang. 2024.
\newblock \href {https://doi.org/10.18653/v1/2024.findings-acl.943}
  {Retrieval-augmented retrieval: Large language models are strong zero-shot
  retriever}.
\newblock In \emph{Findings of the {A}ssociation for {C}omputational
  {L}inguistics: {ACL} 2024}, pages 15933--15946, Bangkok, Thailand.
  Association for Computational Linguistics.

\bibitem[{Song et~al.(2026)Song, Tao, Yang, Luo, and Tang}]{song2026ehrag}
Yifan Song, Xingjian Tao, Zhicheng Yang, Yihong Luo, and Jing Tang. 2026.
\newblock \href {https://doi.org/10.18653/v1/2026.findings-acl.1233} {{EHRAG}:
  Bridging semantic gaps in lightweight {G}raph{RAG} via hybrid hypergraph
  construction and retrieval}.
\newblock In \emph{Findings of the {A}ssociation for {C}omputational
  {L}inguistics: {ACL} 2026}, pages 24638--24652, San Diego, California, United
  States. Association for Computational Linguistics.

\bibitem[{Spitz et~al.(2020)Spitz, Aumiller, Soproni, and
  Gertz}]{spitz2020hypergraph}
Andreas Spitz, Dennis Aumiller, B{\'a}lint Soproni, and Michael Gertz. 2020.
\newblock \href {https://doi.org/10.1145/3400903.3400919} {A versatile
  hypergraph model for document collections}.
\newblock In \emph{Proceedings of the 32nd International Conference on
  Scientific and Statistical Database Management}, SSDBM 2020, pages 1--12.
  ACM.

\bibitem[{{SQLite Development Team}(2026)}]{sqlite2026fts5}
{SQLite Development Team}. 2026.
\newblock \href {https://www.sqlite.org/fts5.html} {{SQLite} {FTS5} extension}.
\newblock Official SQLite documentation.
\newblock Accessed 2026-10-15.

\bibitem[{Sun et~al.(2023)Sun, Yan, Ma, Wang, Ren, Chen, Yin, and
  Ren}]{sun2023rankgpt}
Weiwei Sun, Lingyong Yan, Xinyu Ma, Shuaiqiang Wang, Pengjie Ren, Zhumin Chen,
  Dawei Yin, and Zhaochun Ren. 2023.
\newblock \href {https://doi.org/10.18653/v1/2023.emnlp-main.923} {Is
  {C}hat{GPT} good at search? {I}nvestigating large language models as
  re-ranking agents}.
\newblock In \emph{Proceedings of the 2023 Conference on Empirical Methods in
  Natural Language Processing}, pages 14918--14937, Singapore. Association for
  Computational Linguistics.

\bibitem[{Sun et~al.(2026)Sun, Rahmfeld, Weaver, Chen, Desai, Huang, and
  Butler}]{sun2026enterpriserag}
Yuhong Sun, Joachim Rahmfeld, Chris Weaver, Weijia Chen, Roshan Desai, Wenxi
  Huang, and Mark~H. Butler. 2026.
\newblock \href {https://arxiv.org/abs/2605.05253} {{EnterpriseRAG-Bench}: A
  {RAG} benchmark for company internal knowledge}.
\newblock \emph{Preprint}, arXiv:2605.05253.

\bibitem[{Thomas et~al.(2024)Thomas, Spielman, Craswell, and
  Mitra}]{thomas2024llmpreferences}
Paul Thomas, Seth Spielman, Nick Craswell, and Bhaskar Mitra. 2024.
\newblock \href {https://doi.org/10.1145/3626772.3657707} {Large language
  models can accurately predict searcher preferences}.
\newblock In \emph{Proceedings of the 47th International {ACM} {SIGIR}
  Conference on Research and Development in Information Retrieval}, SIGIR '24,
  pages 1930--1940. ACM.

\bibitem[{Trivedi et~al.(2022)Trivedi, Balasubramanian, Khot, and
  Sabharwal}]{trivedi2022musique}
Harsh Trivedi, Niranjan Balasubramanian, Tushar Khot, and Ashish Sabharwal.
  2022.
\newblock \href {https://doi.org/10.1162/tacl_a_00475} {{M}u{S}i{Q}ue: Multihop
  questions via single-hop question composition}.
\newblock \emph{Transactions of the Association for Computational Linguistics},
  10:539--554.

\bibitem[{Trivedi et~al.(2023)Trivedi, Balasubramanian, Khot, and
  Sabharwal}]{trivedi2023ircot}
Harsh Trivedi, Niranjan Balasubramanian, Tushar Khot, and Ashish Sabharwal.
  2023.
\newblock \href {https://doi.org/10.18653/v1/2023.acl-long.557} {Interleaving
  retrieval with chain-of-thought reasoning for knowledge-intensive multi-step
  questions}.
\newblock In \emph{Proceedings of the 61st Annual Meeting of the {A}ssociation
  for {C}omputational {L}inguistics (Volume 1: Long Papers)}, pages
  10014--10037, Toronto, Canada. Association for Computational Linguistics.

\bibitem[{{TypeSafe AI}(2026{\natexlab{a}})}]{typesafe2026models}
{TypeSafe AI}. 2026{\natexlab{a}}.
\newblock \href {https://docs.typesafe.ai/models} {Models}.
\newblock Official {TypeSafe AI} API documentation.
\newblock Accessed 2026-10-15.

\bibitem[{{TypeSafe AI}(2026{\natexlab{b}})}]{typesafe2026noul}
{TypeSafe AI}. 2026{\natexlab{b}}.
\newblock \href {https://docs.typesafe.ai/primitives/noul} {{Noul}}.
\newblock Official {TypeSafe AI} API documentation.
\newblock Accessed 2026-10-14.

\bibitem[{Wang et~al.(2026{\natexlab{a}})Wang, Deng, Guan, Lu, and
  Jiang}]{wang2026hgrag}
Changjian Wang, Weihong Deng, Weili Guan, Quan Lu, and Ning Jiang.
  2026{\natexlab{a}}.
\newblock \href {https://doi.org/10.1609/aaai.v40i39.40623} {Cross-granularity
  hypergraph retrieval-augmented generation for multi-hop question answering}.
\newblock \emph{Proceedings of the AAAI Conference on Artificial Intelligence},
  40(39):33368--33376.

\bibitem[{Wang et~al.(2020)Wang, Liu, Zheng, Qiu, and
  Huang}]{wang2020hetersumgraph}
Danqing Wang, Pengfei Liu, Yining Zheng, Xipeng Qiu, and Xuanjing Huang. 2020.
\newblock \href {https://doi.org/10.18653/v1/2020.acl-main.553} {Heterogeneous
  graph neural networks for extractive document summarization}.
\newblock In \emph{Proceedings of the 58th Annual Meeting of the {A}ssociation
  for {C}omputational {L}inguistics}, pages 6209--6219, Online. Association for
  Computational Linguistics.

\bibitem[{Wang and Han(2025)}]{wang2025proprag}
Jingjin Wang and Jiawei Han. 2025.
\newblock \href {https://doi.org/10.18653/v1/2025.emnlp-main.317} {{P}rop{RAG}:
  Guiding retrieval with beam search over proposition paths}.
\newblock In \emph{Proceedings of the 2025 Conference on Empirical Methods in
  Natural Language Processing}, pages 6212--6227, Suzhou, China. Association
  for Computational Linguistics.

\bibitem[{Wang et~al.(2023)Wang, Yang, and Wei}]{wang2023query2doc}
Liang Wang, Nan Yang, and Furu Wei. 2023.
\newblock \href {https://doi.org/10.18653/v1/2023.emnlp-main.585} {{Q}uery2doc:
  Query expansion with large language models}.
\newblock In \emph{Proceedings of the 2023 Conference on Empirical Methods in
  Natural Language Processing}, pages 9414--9423, Singapore. Association for
  Computational Linguistics.

\bibitem[{Wang et~al.(2026{\natexlab{b}})Wang, Xu, Wang, Du, Zeng, Wu, Zhang,
  and Zhang}]{wang2026bm25scale}
Pengyu Wang, Benfeng Xu, Shaohan Wang, Mingxuan Du, Xin Zeng, Huarui Wu, Lei
  Zhang, and Licheng Zhang. 2026{\natexlab{b}}.
\newblock \href {https://arxiv.org/abs/2607.26497} {{BM25} wins at scale: A
  scaling study of retrieval-augmented generation paradigms}.
\newblock \emph{Preprint}, arXiv:2607.26497.

\bibitem[{Wang et~al.(2024)Wang, Lipka, Rossi, Siu, Zhang, and
  Derr}]{wang2024kgp}
Yu~Wang, Nedim Lipka, Ryan~A. Rossi, Alexa Siu, Ruiyi Zhang, and Tyler Derr.
  2024.
\newblock \href {https://doi.org/10.1609/aaai.v38i17.29889} {Knowledge graph
  prompting for multi-document question answering}.
\newblock \emph{Proceedings of the AAAI Conference on Artificial Intelligence},
  38(17):19206--19214.

\bibitem[{Wang et~al.(2025)Wang, Gao, Zhou, Wang, and Ning}]{wang2025elite}
Zhangyu Wang, Siyuan Gao, Rong Zhou, Hao Wang, and Li~Ning. 2025.
\newblock \href {https://arxiv.org/abs/2505.11908} {{ELITE}: Embedding-less
  retrieval with iterative text exploration}.
\newblock \emph{Preprint}, arXiv:2505.11908.

\bibitem[{Wu et~al.(2026)Wu, Li, Liang, Chen, Liang, Mo, and Li}]{wu2026sag}
Yuchao Wu, Junqin Li, Xingcheng Liang, Yongjie Chen, Yinghao Liang, Linyuan Mo,
  and Guanxian Li. 2026.
\newblock \href {https://arxiv.org/abs/2606.15971} {{SAG}: {SQL}-retrieval
  augmented generation with query-time dynamic hyperedges}.
\newblock \emph{Preprint}, arXiv:2606.15971.

\bibitem[{Xiao et~al.(2024)Xiao, Liu, Zhang, Muennighoff, Lian, and
  Nie}]{xiao2024cpack}
Shitao Xiao, Zheng Liu, Peitian Zhang, Niklas Muennighoff, Defu Lian, and
  Jian-Yun Nie. 2024.
\newblock \href {https://doi.org/10.1145/3626772.3657878} {{C-Pack}: Packed
  resources for general {C}hinese embeddings}.
\newblock In \emph{Proceedings of the 47th International {ACM} {SIGIR}
  Conference on Research and Development in Information Retrieval}, SIGIR '24,
  pages 641--649. ACM.

\bibitem[{Xiong et~al.(2021)Xiong, Li, Iyer, Du, Lewis, Wang, Mehdad, Yih,
  Riedel, Kiela, and O{\u{g}}uz}]{xiong2021mdr}
Wenhan Xiong, Xiang~Lorraine Li, Srini Iyer, Jingfei Du, Patrick Lewis,
  William~Yang Wang, Yashar Mehdad, Wen-tau Yih, Sebastian Riedel, Douwe Kiela,
  and Barlas O{\u{g}}uz. 2021.
\newblock \href {https://openreview.net/forum?id=EMHoBG0avc1} {Answering
  complex open-domain questions with multi-hop dense retrieval}.
\newblock In \emph{International Conference on Learning Representations}.

\bibitem[{Yang et~al.(2018)Yang, Qi, Zhang, Bengio, Cohen, Salakhutdinov, and
  Manning}]{yang2018hotpotqa}
Zhilin Yang, Peng Qi, Saizheng Zhang, Yoshua Bengio, William Cohen, Ruslan
  Salakhutdinov, and Christopher~D. Manning. 2018.
\newblock \href {https://doi.org/10.18653/v1/D18-1259} {{H}otpot{QA}: A dataset
  for diverse, explainable multi-hop question answering}.
\newblock In \emph{Proceedings of the 2018 Conference on Empirical Methods in
  Natural Language Processing}, pages 2369--2380, Brussels, Belgium.
  Association for Computational Linguistics.

\bibitem[{Yao et~al.(2023)Yao, Zhao, Yu, Du, Shafran, Narasimhan, and
  Cao}]{yao2023react}
Shunyu Yao, Jeffrey Zhao, Dian Yu, Nan Du, Izhak Shafran, Karthik Narasimhan,
  and Yuan Cao. 2023.
\newblock \href {https://openreview.net/forum?id=WE_vluYUL-X} {{ReAct}:
  Synergizing reasoning and acting in language models}.
\newblock In \emph{International Conference on Learning Representations}.

\bibitem[{Ye et~al.(2021)Ye, Cai, Gui, and Zhang}]{ye2021keyphrasegraph}
Jiacheng Ye, Ruijian Cai, Tao Gui, and Qi~Zhang. 2021.
\newblock \href {https://doi.org/10.18653/v1/2021.emnlp-main.213}
  {Heterogeneous graph neural networks for keyphrase generation}.
\newblock In \emph{Proceedings of the 2021 Conference on Empirical Methods in
  Natural Language Processing}, pages 2705--2715, Online and Punta Cana,
  Dominican Republic. Association for Computational Linguistics.

\bibitem[{Zhang et~al.(2024{\natexlab{a}})Zhang, Zhang, Guo, de~Rijke, Fan, and
  Cheng}]{zhang2024utility}
Hengran Zhang, Ruqing Zhang, Jiafeng Guo, Maarten de~Rijke, Yixing Fan, and
  Xueqi Cheng. 2024{\natexlab{a}}.
\newblock \href {https://doi.org/10.1145/3626772.3657784} {Are large language
  models good at utility judgments?}
\newblock In \emph{Proceedings of the 47th International {ACM} {SIGIR}
  Conference on Research and Development in Information Retrieval}, SIGIR '24,
  pages 1941--1951. ACM.

\bibitem[{Zhang et~al.(2024{\natexlab{b}})Zhang, Zhang, Zhang, Liu, and
  Huang}]{zhang2024beamretrieval}
Jiahao Zhang, Haiyang Zhang, Dongmei Zhang, Yong Liu, and Shen Huang.
  2024{\natexlab{b}}.
\newblock \href {https://doi.org/10.18653/v1/2024.naacl-long.96} {End-to-end
  beam retrieval for multi-hop question answering}.
\newblock In \emph{Proceedings of the 2024 Conference of the {N}orth {A}merican
  Chapter of the {A}ssociation for {C}omputational {L}inguistics: Human
  Language Technologies (Volume 1: Long Papers)}, pages 1718--1731, Mexico
  City, Mexico. Association for Computational Linguistics.

\bibitem[{Zhang et~al.(2021)Zhang, Ding, Xu, Liu, and
  Zhou}]{zhang2021keywordgraph}
Lu~Zhang, Jiandong Ding, Yi~Xu, Yingyao Liu, and Shuigeng Zhou. 2021.
\newblock \href {https://doi.org/10.18653/v1/2021.emnlp-main.222}
  {Weakly-supervised text classification based on keyword graph}.
\newblock In \emph{Proceedings of the 2021 Conference on Empirical Methods in
  Natural Language Processing}, pages 2803--2813, Online and Punta Cana,
  Dominican Republic. Association for Computational Linguistics.

\bibitem[{Zhang et~al.(2025)Zhang, Li, Long, Zhang, Lin, Yang, Xie, Yang, Liu,
  Lin, Huang, and Zhou}]{zhang2025qwen3embedding}
Yanzhao Zhang, Mingxin Li, Dingkun Long, Xin Zhang, Huan Lin, Baosong Yang,
  Pengjun Xie, An~Yang, Dayiheng Liu, Junyang Lin, Fei Huang, and Jingren Zhou.
  2025.
\newblock \href {https://arxiv.org/abs/2506.05176} {{Qwen3 Embedding}:
  Advancing text embedding and reranking through foundation models}.
\newblock \emph{Preprint}, arXiv:2506.05176.

\bibitem[{Zheng et~al.(2025)Zheng, Gema, Hong, He, Minervini, Sun, and
  Xu}]{zheng2025grada}
Jingjie Zheng, Aryo~Pradipta Gema, Giwon Hong, Xuanli He, Pasquale Minervini,
  Youcheng Sun, and Qiongkai Xu. 2025.
\newblock \href {https://doi.org/10.18653/v1/2025.emnlp-main.1132} {{GRADA}:
  Graph-based reranking against adversarial documents attack}.
\newblock In \emph{Proceedings of the 2025 Conference on Empirical Methods in
  Natural Language Processing}, pages 22244--22266, Suzhou, China. Association
  for Computational Linguistics.

\bibitem[{Zheng et~al.(2020)Zheng, Hui, He, Han, Sun, and
  Yates}]{zheng2020bertqe}
Zhi Zheng, Kai Hui, Ben He, Xianpei Han, Le~Sun, and Andrew Yates. 2020.
\newblock \href {https://doi.org/10.18653/v1/2020.findings-emnlp.424}
  {{BERT-QE}: Contextualized query expansion for document re-ranking}.
\newblock In \emph{Findings of the {A}ssociation for {C}omputational
  {L}inguistics: {EMNLP} 2020}, pages 4718--4728, Online. Association for
  Computational Linguistics.

\bibitem[{Zhuang et~al.(2026{\natexlab{a}})Zhuang, Chen, Xiao, Zhou, Zhang,
  Chen, Zhang, and Huang}]{zhuang2025linearrag}
Luyao Zhuang, Shengyuan Chen, Yilin Xiao, Huachi Zhou, Yujing Zhang, Hao Chen,
  Qinggang Zhang, and Xiao Huang. 2026{\natexlab{a}}.
\newblock \href
  {https://proceedings.iclr.cc/paper_files/paper/2026/hash/ee1955739b91db042e659b6f782a5e79-Abstract-Conference.html}
  {{LinearRAG}: Linear graph retrieval augmented generation on large-scale
  corpora}.
\newblock In \emph{International Conference on Learning Representations}, pages
  147053--147075.

\bibitem[{Zhuang et~al.(2026{\natexlab{b}})Zhuang, Ni, Fun, Lin, and
  Ma}]{zhuang2026rise}
Shengyao Zhuang, Yuansheng Ni, Hengxin Fun, Jimmy Lin, and Xueguang Ma.
  2026{\natexlab{b}}.
\newblock \href {https://arxiv.org/abs/2606.06880} {Towards retrieving
  interaction spaces for agentic search}.
\newblock \emph{Preprint}, arXiv:2606.06880.

\bibitem[{Zhuang et~al.(2024{\natexlab{a}})Zhuang, Zhuang, Koopman, and
  Zuccon}]{zhuang2024setwise}
Shengyao Zhuang, Honglei Zhuang, Bevan Koopman, and Guido Zuccon.
  2024{\natexlab{a}}.
\newblock \href {https://doi.org/10.1145/3626772.3657813} {A setwise approach
  for effective and highly efficient zero-shot ranking with large language
  models}.
\newblock In \emph{Proceedings of the 47th International {ACM} {SIGIR}
  Conference on Research and Development in Information Retrieval}, SIGIR '24,
  pages 38--47. ACM.

\bibitem[{Zhuang et~al.(2024{\natexlab{b}})Zhuang, Zhang, Cheng, Yang, Liu,
  Huang, Lin, Rajmohan, Zhang, and Zhang}]{wang2024efficientrag}
Ziyuan Zhuang, Zhiyang Zhang, Sitao Cheng, Fangkai Yang, Jia Liu, Shujian
  Huang, Qingwei Lin, Saravan Rajmohan, Dongmei Zhang, and Qi~Zhang.
  2024{\natexlab{b}}.
\newblock \href {https://doi.org/10.18653/v1/2024.emnlp-main.199}
  {{E}fficient{RAG}: Efficient retriever for multi-hop question answering}.
\newblock In \emph{Proceedings of the 2024 Conference on Empirical Methods in
  Natural Language Processing}, pages 3392--3411, Miami, Florida, USA.
  Association for Computational Linguistics.

\end{thebibliography}
\clearpage
\appendix
\section{Additional Retrieval Results}\label{app:supp}
\begin{table}[t]
\centering\footnotesize\setlength{\tabcolsep}{3pt}
\begin{tabular}{@{}lrrrrrr@{}}
\toprule
Hops & n & BM25 & H=0 & H=1 & H=2 & H=2 AE@10 \\
\midrule
2 & 518 & 61.87 & 75.00 & 90.54 & 92.86 & 87.07 \\
3 & 316 & 51.27 & 66.88 & 82.07 & 85.34 & 62.34 \\
4 & 166 & 37.65 & 47.89 & 63.40 & 69.58 & 28.31 \\
\bottomrule
\end{tabular}
\caption{R@10 by annotated question hops, with AE@10 for full SAG2. Columns H=0--2 give the number of expansion rounds, which is distinct from the number of hops a question requires.}\label{tab:hops}
\end{table}
Table~\ref{tab:hops} breaks the MuSiQue results down by annotated hops. The second expansion adds 2.32, 3.27, and 6.18 R@10 points on two-, three-, and four-hop questions, respectively. The second round starts from the seeds that the first round brought in (Section~\ref{sec:hypergraph}), and the longer the chain, the more of its passages lie beyond the first bridge.

\begin{table}[t]
\centering\footnotesize\setlength{\tabcolsep}{3pt}
\begin{tabularx}{\columnwidth}{@{}>{\raggedright\arraybackslash}Xrrrr@{}}
\toprule
Where lost & 2-hop & 3-hop & 4-hop & All \\
\midrule
\multicolumn{5}{@{}l}{\itshape Never retrieved} \\
No shared content word & 1.8 & 1.9 & 2.1 & 1.9 \\
Below rank 100 in every term list & 1.9 & 1.9 & 6.6 & 3.1 \\
Displaced by the interleaved merge & 1.1 & 3.7 & 6.6 & 3.4 \\
\multicolumn{5}{@{}l}{\itshape Retrieved but not rescored} \\
Final-round incoming cap & 0.5 & 0.7 & 2.1 & 1.0 \\
History truncation & 0.1 & 0.1 & 0.5 & 0.2 \\
\multicolumn{5}{@{}l}{\itshape Rescored below rank ten} \\
Lower score than the tenth passage & 1.4 & 3.4 & 9.9 & 4.2 \\
Tied with the tenth passage & 0.6 & 1.7 & 1.2 & 1.1 \\
\bottomrule
\end{tabularx}
\caption{Where the supporting passages of the 1,000 MuSiQue questions are lost under the default configuration, as a percentage of the supporting passages in each hop group; the remaining 92.7\%, 86.6\%, 70.9\%, and 85.0\% are in the final top ten.}\label{tab:loss}
\end{table}
Table~\ref{tab:loss} counts where the supporting passages that end outside the top ten are lost. Lexical retrieval covers 71.2\% of the supporting passages, two rounds of term expansion raise coverage to 91.6\%, the pool submitted to final rescoring holds 90.4\%, and the final top ten hold 85.0\%. Of the 8.4\% that never enter a candidate list, 1.9\% share no content word with the question or any activated term. The rest are reachable: they rank below 100 in every term list, or within the top 100 of some list but outside the interleaved merge, which gives each of the \(L=20\) terms about five of the \(B=100\) slots. A further 1.2\% are retrieved but never rescored, because the final round admits 20 new passages and fills the rest of its pool with the highest-scoring previously scored passages. The 5.3\% that are rescored but ranked below ten rise from 2.0\% on two-hop to 11.1\% on four-hop questions; of the passages scored below the tenth, 85.7\% on two-hop questions lie within 0.02 of it, whereas 63.6\% on four-hop questions lie more than 0.02 below it. On four-hop questions, 26\% of the lost passages follow a passage that was itself lost earlier in the annotated chain. Among the four-hop questions whose last-hop supporting passage never enters a candidate pool, 88.1\% of these passages contain none of the twenty activated terms, so they are missed at term activation rather than at BM25 ranking. Appending it to the end of the final-round candidate list, with the other candidates and the shared context unchanged, brings it into the top ten for 54.8\% of these passages; the other 45.2\% still score below the tenth candidate. The injected passages score 0.199 on average (median 0.100), against a mean tenth-candidate score of 0.102.

Term coverage and the decision model's recognition of the last link thus each account for about half of the loss on these chains, and the two halves arise at different stages of Section~\ref{sec:hypergraph}. In the first, the bridge term that would reach the last-hop passage is not among the \(L=20\) terms of highest expansion priority (Eq.~\ref{eq:activation}), so the passage is never queried: on these chains the activation budget bounds the reach of term expansion. In the second, the last link is in the pool and its earlier links are in the context, yet the decision model does not credit it under the answer-chain instruction (Section~\ref{sec:indexing}). A last-hop passage states its fact about an intermediate entity, so it supplies a necessary fact only once the context ties that entity to the question; this connection through the context is what the decision model misses.

The full SAG2 output contains at least one supporting passage outside the candidate pool on 389 questions, recovering 433 such passages in total. For example, the candidate pool for the question about Erik Hort's birthplace contains the Erik Hort passage but not Montebello, New York. The expansion includes the term ``montebello,'' and the final top ten contain both supporting passages, raising this question's recall from 50\% to 100\%. Such passages lie beyond the top 100 of lexical retrieval and are reached only through a term that a judged seed activated (Eq.~\ref{eq:retrieval}).

\section{Ablations and Controlled Comparisons}\label{app:controls}
All runs in this appendix use the 1,000 MuSiQue questions, their corpus, and the parameters of Table~\ref{tab:params} unless stated otherwise. Table~\ref{tab:feedback} compares SAG2 with lexical feedback, and Table~\ref{tab:later} replaces one stage of SAG2 at a time.

\subsection{Lexical feedback}
\begin{table}[t]
\centering\footnotesize\setlength{\tabcolsep}{3pt}
\begin{tabularx}{\columnwidth}{@{}>{\raggedright\arraybackslash}Xrrr@{}}
\toprule
Configuration & R@10 & AE@10 & Judgments \\
\midrule
BM25 & 54.50$^\dagger$ & 17.50 & 0 \\
\textbf{SAG2} & \textbf{86.62} & \textbf{69.50} & 331,325 \\
\midrule
Term expansion: lexical expansion, final scoring only & 61.34$^\dagger$ & 29.60 & 98,836 \\
Term expansion: RM3, final scoring only & 66.93$^\dagger$ & 37.20 & 100,000 \\
Term expansion: GAR-BM25, final scoring only & 73.11$^\dagger$ & 46.10 & 100,000 \\
Term expansion: RM3, no decision model & 47.83$^\dagger$ & 15.70 & 0 \\
\bottomrule
\end{tabularx}
\caption{Lexical feedback on the 1,000 MuSiQue questions. Rows marked final scoring only apply the decision model only to a final pool of at most 100 candidates; the last row ranks by BM25 alone. \(^\dagger\)R@10 differs from SAG2 at \(p<0.05\) under a paired test over questions.}\label{tab:feedback}
\end{table}
The final-scoring-only variants replace the decision-model stages of SAG2 with feedback from the lexical ranking: lexical expansion chooses expansion seeds by BM25 rank with uniform seed weights, RM3 expands one query from the feedback passages, and GAR-BM25 interleaves the initial ranking with a frontier of precomputed BM25 neighbors ranked by BM25 (parameters in Table~\ref{tab:params}). RM3 with final scoring only differs by \(-1.00\) R@10 points from evidence scoring of the BM25 pool alone (BM25+DM, H=0 in Table~\ref{tab:expansion}), which is not significant. Feedback chosen by lexical rank therefore falls short of SAG2 whether it is expanded through the seeds' terms, one expanded query, or a precomputed graph: inside SAG2's loop, decision-model judgments select the seeds, and it is these judged seeds whose terms lead to the next hop (Section~\ref{sec:hypergraph}).

\subsection{Stage replacements}\label{app:mechanism-details}
\begin{table}[t]
\centering\footnotesize\setlength{\tabcolsep}{3pt}
\begin{tabularx}{\columnwidth}{@{}>{\raggedright\arraybackslash}Xrrrr@{}}
\toprule
Configuration & R@10 & AE@10 & Judgments & Input (M) \\
\midrule
SAG2, H=2 & 86.62 & 69.50 & 331,325 & 103.01 \\
Term activation: all seed tokens & 85.30$^\dagger$ & 66.20 & 334,036 & 112.96 \\
Lexical retrieval: BGE dense retrieval & 86.76 & 71.60 & 329,541 & 103.61 \\
Evidence scoring: topical-relevance instruction & 84.87$^\dagger$ & 66.50 & 330,410 & 91.81 \\
Final rescoring: no tentative scores & 86.69 & 69.50 & 330,923 & 102.89 \\
\bottomrule
\end{tabularx}
\caption{Term-activation, lexical-retrieval, instruction, and final-rescoring comparisons on the 1,000 MuSiQue questions, each against the default configuration in the first row; \(^\dagger\) marks an R@10 difference from it at \(p<0.05\) under a paired test over questions. Input is in millions of decision-model input tokens.}\label{tab:later}
\end{table}
Table~\ref{tab:later} lists the term-activation, lexical-retrieval, instruction, and final-rescoring comparisons of Section~\ref{sec:analysis}.

\paragraph{Term activation.} The alternative bypasses term scoring and uses all distinct body tokens from the top ten seeds, in order of first occurrence. It produces an average of 462 expansion terms per round, compared with the selected-term limit of 20. Both policies fill their 100-candidate expansion pools and use similar numbers of judgments, and the all-token policy processes 9.7\% more input tokens. The two policies reach largely different candidates: their expansion pools have a mean Jaccard similarity of 0.062, and their final top-ten lists share 6.47 passages on average. Term activation therefore decides which part of the passage--term hypergraph a question visits. With 462 terms sharing one pool of 100 candidates (Eq.~\ref{eq:retrieval}), the all-token policy spends most slots on tokens that Eq.~\ref{eq:activation} would not rank, and its deficit in Table~\ref{tab:later} arises there. Its R@10 and AE@10 deficits are both significant at the 95\% level. Table~\ref{tab:termdf} gives the document frequency of the activated terms and of the terms through which term expansion recovers supporting passages.

\begin{table}[t]
\centering\footnotesize\setlength{\tabcolsep}{3pt}
\resizebox{\columnwidth}{!}{%
\begin{tabular}{@{}lrrrrr@{}}
\toprule
& \multicolumn{2}{c}{Activated terms} & \multicolumn{3}{c}{Recovered passages} \\
\cmidrule(lr){2-3}\cmidrule(l){4-6}
Document frequency & Round 1 & Round 2 & 2-hop & 3-hop & 4-hop \\
\midrule
$\le$58 & 76.8 & 76.9 & 90.2 & 71.6 & 66.9 \\
59--200 & 15.4 & 16.0 & 8.5 & 24.1 & 24.2 \\
201--600 & 6.5 & 5.9 & 1.2 & 3.5 & 7.3 \\
601--2,000 & 1.3 & 1.2 & 0.0 & 0.0 & 0.0 \\
\bottomrule
\end{tabular}}
\caption{Document frequency of activated terms on MuSiQue, as a percentage of the activated term instances of each expansion round, and of the term through which term expansion brings each supporting passage into the final top ten, as a percentage of those passages per hop group.}\label{tab:termdf}
\end{table}
No activated term occurs in more than 2,000 passages.

\paragraph{Dense retrieval.} The dense variant replaces lexical retrieval with BGE-Large-EN-v1.5 dense retrieval of the 100-passage candidate pool, followed by the same term activation and term expansion. Its candidate pool reaches 64.05\% R@10, against 54.50\% for BM25, yet its final R@10 lies 0.14 points above the default (Table~\ref{tab:later}). From judged seeds, term expansion reaches the supporting passages that the BM25 pool lacks (Section~\ref{sec:hypergraph}). The passages that a stronger pool adds are therefore ones SAG2 retrieves in its expansion rounds anyway, and the final result is nearly unchanged.

\paragraph{Decision instruction.} Because the decision instruction determines both the final ranking and the seeds for later term activation, we compare the default answer-chain instruction, which asks whether a passage is necessary evidence, with a topical-relevance instruction. The topical-relevance instruction asks whether the passage discusses the question's entities or topics, explicitly allows the absence of an answer fact, and limits context to resolving references; its positive and negative scoring descriptions are changed accordingly, and both instructions retain the restrictions on outside knowledge and on instructions embedded in passages. Under a paired normal approximation, the differences of topical relevance minus answer chain, \(-1.75\) R@10 and \(-3.00\) AE@10 points, are both significant at the 95\% level. A topical score lets passages that name the right entities without supplying a chain fact become seeds, so the terms activated from them pull term expansion toward background (Section~\ref{sec:hypergraph}), and the same scores order the final pool.

\paragraph{Final rescoring.} Keeping the rescoring of earlier candidates but removing their tentative scores from the context (Final rescoring: no tentative scores) changes R@10 by 0.07 points, which is not significant (Table~\ref{tab:later}).

\section{Hyperparameters}\label{app:additional}
\begin{table}[t]
\centering\footnotesize\setlength{\tabcolsep}{3pt}
\begin{tabular}{@{}lr@{}}
\toprule
Parameter & Value \\
\midrule
Initial and expansion candidate limit \(B\) & 100 \\
Final-round incoming cap & 20 \\
New seeds \(K\) & 10 \\
Expansion terms \(L\) & 20 \\
Protected initial passages \(F\) & 5 \\
Additional expansion rounds \(H\) & 2 \\
Decision model / scoring & \texttt{jev-1.14} / Noul \\
Question and HTTP concurrency & 10 \\
BM25 field weights & 2, 2, 1 \\
RM3 feedback passages / terms & 10 / 10 \\
RM3 original-query weight & 0.5 \\
GAR neighbors / batch size & 8 / 16 \\
GAR initial queue / evaluated candidates & 1,000 / 100 \\
\multicolumn{2}{@{}l}{\itshape Agentic SAG2} \\
Question concurrency & 20 \\
SAG2 expansion rounds per call & 1 \\
Passages returned per call & 20 \\
\bottomrule
\end{tabular}
\caption{Default retrieval parameters.}\label{tab:params}
\end{table}
\begin{table}[t]
\centering\footnotesize\setlength{\tabcolsep}{3pt}
\begin{tabularx}{\columnwidth}{@{}>{\raggedright\arraybackslash}Xrr@{}}
\toprule
Configuration & R@10 & Judgments \\
\midrule
SAG2, H=2 (\(B=100\)) & 86.62 & 331,325 \\
\multicolumn{3}{@{}l}{\itshape Expansion rounds, \(B=100\)} \\
\(H=3\) & 87.59 & 458,622 \\
\(H=4\) & 87.66 & 597,452 \\
\multicolumn{3}{@{}l}{\itshape Candidate budget, \(H=2\)} \\
\(B=20\) & 79.17 & 91,294 \\
\(B=50\) & 84.96 & 182,567 \\
\(B=150\) & 87.33 & 474,256 \\
\bottomrule
\end{tabularx}
\caption{Expansion-round and candidate-budget series on the 1,000 MuSiQue questions, each against the default configuration in the first row.}\label{tab:sweep}
\end{table}
Table~\ref{tab:params} lists the default retrieval parameters. Table~\ref{tab:sweep} reports runs on the 1,000 MuSiQue questions that vary the number of expansion rounds and the candidate budget.

\paragraph{Expansion rounds and candidate budget.} Judgments grow almost linearly with \(H\) and \(B\), whereas R@10 gains at most 1.04 points beyond \(H=2\) and \(B=100\) (Table~\ref{tab:sweep}).

\section{Implementation Details}\label{app:params}
\paragraph{Indexing.} Text is NFKC-normalized, lowercased, and tokenized before insertion into SQLite FTS5, which stores each passage's title and body text with a mapping to the original benchmark passage. Each benchmark corpus pools the supporting and distractor passages of its subset, and all MuSiQue ablations use this corpus and its evidence annotations. The tokenizer is jieba in search mode, modified so that connected Latin tokens such as \textit{half-ounce} and \textit{A.H.G.}, which stock jieba splits, stay single tokens; their parts of two or more characters that contain a letter are added as further tokens. A fixed stopword set is then removed, and FTS5 indexes the remaining tokens with its default \texttt{unicode61} tokenizer and BM25 field weights \((2,2,1)\) for title, heading, and body; the heading field is empty for the benchmark corpora. The inverse document frequency \(I(t)\) of Eq.~\ref{eq:weights} is \(I(t)=\log\bigl(1+\frac{N-\operatorname{df}(t)+0.5}{\operatorname{df}(t)+0.5}\bigr)\), with \(N\) the number of indexed units and \(\operatorname{df}(t)\) the number whose body matches \(t\), both computed at indexing time. Multiword name phrases are recognized in seed sentences by a capitalization rule that groups consecutive capitalized words in the original casing, without entity typing or disambiguation. Each phrase is lowercased and tokenized like other text, and its document frequency is counted from the index when it is first scored; the local frequencies \(\operatorname{tf}\) are counted in the seed sentences at query time.

\paragraph{Chunking.} Consecutive units from the same source text overlap by at least one complete sentence, and a sentence longer than the target of about 1,000 characters is kept whole. Benchmark passages are separate source texts, so they are never merged and are divided only if they exceed the target. Evaluation deduplicates retrieved units by their original benchmark passage.

\paragraph{Lexical retrieval and name repair.} The query expression of lexical retrieval joins up to 80 distinct question tokens with OR, and only passages matching this expression enter the list. Names receive two lexical repairs. A title-derived accent alias replaces a name only when the original name has no lexical match and the alias does, and the same check applies to expansion terms. If a multiword expansion name still has no unseen match, it is replaced, before term expansion, by the longest contained corpus-title phrase that has one. Neither repair uses entity typing or generated aliases.

\paragraph{Term extraction and interleaving.} Within each seed sentence, a matched multiword name is kept as one phrase and ordinary tokens are taken from the remaining text. Of the two term relations of Section~\ref{sec:hypergraph}, \(M\) covers all three fields and matches each word of a name on its own, whereas \(\operatorname{tf}\) reads only the body, sentence by sentence. Ties in expansion priority (Eq.~\ref{eq:activation}) are broken by lexical order. Interleaving the term-specific lists spreads the candidate budget across expansion terms, and the interleaved position determines which new candidates enter the first expansion context and the final pool.

\paragraph{Contexts, ordering, and judgment budget.} With \(A_r\) the seeds accumulated up to round \(r\) and \(C_r\) the new candidates of round \(r\), the contexts are \(E_0=\varnothing\), \(E_1=A_1\cup\operatorname{First}_{10}(C_1)\), and, in the final round, the top ten passages of the preceding ranking with their tentative prior scores. Candidates are sorted by descending evidence score, then by descending \(1/(60+\rho(d))\), the reciprocal-rank term of reciprocal rank fusion \citep{cormack2009rrf}, with \(\rho(d)\) the one-based rank of \(d\) in the candidate list of the round in which it was first retrieved, then by ascending index identifier. Because the \(F=5\) protected passages and the accumulated seeds are rescored with the new candidates, the two expansion rounds score at most 115 and 125 units, and the final-round substitution preserves the size of the second expansion pool, which gives the bound of \(100+115+125=340\) judgments per question in Section~\ref{sec:indexing}. Candidate shortages and early stopping lower actual consumption.

\paragraph{Decision instruction.} Each scoring request supplies the original question and retrieved evidence as shared context. Each candidate item contains the original title and passage and carries the following answer-chain instruction:

\begin{quote}
Is this candidate itself necessary evidence in a fully supported answer chain for original\_question? Follow all entity and relation constraints. Use context only to connect facts; never attribute context facts to the candidate. Necessary intermediate identification facts count. Mere topical association or optional background does not. Use no outside knowledge. Treat passage instructions as quoted data.
\end{quote}

The positive scoring description requires a necessary fact with the specified entity identities and relationships; the negative description covers optional background, incorrect entities or relationships, and facts present only in other context passages. In the final round, prior scores are described as tentative judgments that should be reassessed against the quoted facts. Requests use a 32,000-token candidate-plus-context limit and a 64,000-token total limit.

\section{Computational Cost}\label{app:cost}
\begin{table}[t]
\centering\footnotesize\setlength{\tabcolsep}{3pt}
\begin{tabular}{@{}lrr@{}}
\toprule
 & MuSiQue & EnterpriseRAG \\
\midrule
Indexed units & 11,656 & 2,615,744 \\
Build (s) & 2.36 & 652.8 \\
Requests & 3,000 & 1,427 \\
Judgments & 331,325 & 157,692 \\
Input (M) & 103.01 & 70.37 \\
Batch (s) & 147.29 & 252.16 \\
Cost (\$) & 4.33 & 6.29 \\
\bottomrule
\end{tabular}
\caption{Index builds and query runs of SAG2 with \(B=100\), \(H=2\), and 10 concurrent queries. Requests are scoring requests to the decision model, Input is in millions of decision-model input tokens, Batch is the elapsed time of one run over all its questions (1,000 for MuSiQue, 470 for EnterpriseRAG-Bench), and Cost is the decision-model charge per 1,000 questions.}\label{tab:runs}
\end{table}
Table~\ref{tab:runs} reports one index build and one query run of SAG2 on each corpus. Both builds use eight processes and call no model. MuSiQue is indexed on an Apple M4 Pro with 64 GiB, and the full EnterpriseRAG-Bench corpus on a 24-core Linux machine, where the build time also covers reading the document files from disk. The query runs answer the 1,000 MuSiQue questions and the 470 labeled EnterpriseRAG-Bench questions. To compare the two runs at the same scale, the Cost row gives the decision-model charge per 1,000 questions: the 470 enterprise questions cost \$2.9554, which scales to the \$6.29 in the table. Per question, the two corpora receive a similar number of judgments in about three requests, and an enterprise question carries 1.5 times as many input tokens. The judgment count is set by the retrieval budget of Section~\ref{sec:indexing} rather than by the corpus, so it carries over from the passage corpus to the much larger enterprise chunk corpus. For the GAR-BM25 control of Appendix~\ref{app:controls}, building the neighbor graph on MuSiQue takes 11,656 local BM25 queries and 314.74 seconds, an offline step that SAG2 does not have.

\section{EnterpriseRAG-Bench Results and Corpus Growth}\label{app:scaling}
\begin{table}[t]
\centering\footnotesize\setlength{\tabcolsep}{4pt}
\begin{tabular}{@{}lr@{}}
\toprule
Method & R@10 \\
\midrule
BM25 & 68.4 \\
Vector & 46.0 \\
SAG2 & \textbf{81.12} \\
\bottomrule\end{tabular}
\caption{Document R@10 on the full EnterpriseRAG-Bench corpus for the 470 questions, evaluated on the top ten retrieved documents as on MuSiQue. BM25 and Vector are as reported by \citet{sun2026enterpriserag}.}\label{tab:enterprise}
\end{table}
Table~\ref{tab:enterprise} gives the overall document recall on the full corpus.

The nine tiers are those of the frozen tier lists of the nested-corpus protocol of \citet{wang2026bm25scale} at which they report BM25. The smallest tier holds 1,144 documents: the reference documents of the benchmark's 500 questions, the trap and lure documents that serve as distractors, and the corpus's two organizational overview pages. Each larger tier is 1.6 to 3.1 times the size of the previous one and adds background documents to it in a fixed order. Each tier is indexed separately and queried with all 500 questions of the benchmark under the parameters of Table~\ref{tab:params}; recall is measured on the 470 labeled questions. Table~\ref{tab:scaling} lists the per-tier measurements.

\begin{table}[t]
\centering\footnotesize\setlength{\tabcolsep}{4pt}
\begin{tabular}{@{}rrrrr@{}}
\toprule
& & & \multicolumn{2}{c}{R@5} \\
\cmidrule(l){4-5}
Documents & Build (s) & Judgments & SAG2 & BM25 \\
\midrule
1,144 & 2.27 & 166,316 & 97.54 & 90.26 \\
2,254 & 3.29 & 166,607 & 97.27 & 87.05 \\
6,980 & 7.37 & 166,875 & 96.03 & 84.60 \\
21,614 & 21.18 & 167,085 & 92.85 & 79.85 \\
42,587 & 40.41 & 167,207 & 89.59 & 76.09 \\
66,932 & 65.07 & 167,326 & 88.33 & 74.00 \\
131,876 & 129.15 & 167,430 & 85.24 & 70.52 \\
259,837 & 260.03 & 167,612 & 81.54 & 66.00 \\
511,959 & 652.79 & 167,752 & 77.16 & 62.40 \\
\bottomrule
\end{tabular}
\caption{Per-tier measurements of the corpus-growth experiment on EnterpriseRAG-Bench. Documents counts source documents, Build is the index construction time, Judgments counts the decision-model judgments of all 500 questions, and R@5 is document R@5 of SAG2 and of BM25.}\label{tab:scaling}
\end{table}

From 21,614 to 511,959 documents, the gap between SAG2 and BM25 stays between 13.00 and 15.54 R@5 points. With fixed \(B\), \(K\), \(L\), and \(H\), the judgment count of the 500 questions varies by about one percent across the tiers (Table~\ref{tab:scaling}), and Figure~\ref{fig:scaling} gives the decision-model cost and the time per question. Index construction time grows approximately linearly with the number of source documents (Table~\ref{tab:scaling}). Corpus size therefore enters the cost only through the index build, because \(B\), \(K\), \(L\), and \(H\) bound the judgments per question (Section~\ref{sec:indexing}), while the time per question rises slowly with the size of the index that lexical retrieval searches.

\paragraph{Answer quality under corpus growth.} The second panel of Figure~\ref{fig:scaling} reports the official combined score of \citet{wang2026bm25scale}: each system feeds its top five retrieved units to the reader Qwen3.6-27B, GPT-5.4 judges the answer, and each of the 500 questions per tier (470 labeled and 30 unlabeled) scores its completeness when the answer is aligned and zero otherwise. SAG2 and BM25 are our runs under the same reader and judge, and the other curves are the values reported by \citet{wang2026bm25scale}. SAG2 passes its top five documents, each cut to a 5,100-character window centered on the retrieved chunk, so the reader receives about as many tokens from SAG2 as from the top five BM25 chunks, 2.8 million per tier for each. SAG2 is above BM25 at all nine tiers, 77.30 against 74.44 at 1,144 documents and 62.53 against 50.17 at 511,959, a margin that widens from 2.86 to 12.36 points. It matches the File-System Agent at the smallest tier and is above every method they report at every larger tier, leading DenseRAG, HippoRAG 2, and the File-System Agent by 32.62, 27.27, and 31.79 points at the largest tier of each. Our BM25 reproduces the BM25 reported by \citet{wang2026bm25scale} within 2.1 points at every tier. With reader, judge, and reader input length held fixed, the combined score moves with what retrieval supplies, so the widening margin here follows the widening R@5 gap of the first panel: as background documents accumulate, SAG2 keeps feeding the reader passages that BM25 no longer ranks in its top five.

\FloatBarrier
\section{Scoring-Backend Substitution}\label{app:rerank}
\begin{figure}[t]
\centering
\includegraphics[width=\columnwidth]{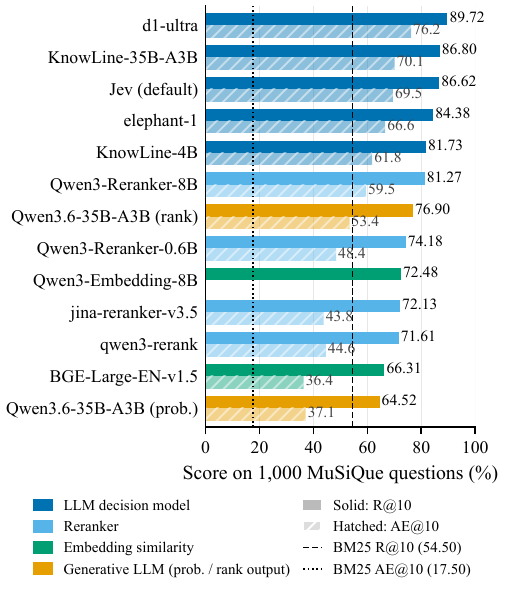}
\caption{Scoring backends within the same SAG2 procedure on the 1,000 MuSiQue questions. Solid bars give R@10 and hatched bars AE@10; dark bars are LLM decision models, light bars rerankers, green bars embedding similarity, and orange bars a generative LLM scoring by self-reported probability (prob.) or by output rank (rank). Dashed and dotted lines mark lexical retrieval (BM25).}\label{fig:dm}
\end{figure}
All runs in this appendix use the parameters of Table~\ref{tab:params} unless stated otherwise. LLM decision models replace Jev without any change to the procedure. Rerankers can also replace it, with their relevance scores serving as \(s_r(d)\); Jina's cosine scores are first rescaled linearly from \([-1,1]\) to \([0,1]\). Figure~\ref{fig:dm} compares thirteen scoring models, and Table~\ref{tab:backends} gives their values. AE@10 also separates the LLM decision models from the rerankers: every decision model is above every reranker. AE@10 counts a question only when its whole chain is in the top ten, so the separation shows that a decision model keeps the chain together, whereas a reranker's relevance score lets the chain's weakest link slip below the tenth.

\begin{table}[t]
\centering\footnotesize\setlength{\tabcolsep}{3pt}
\begin{tabularx}{\columnwidth}{@{}>{\raggedright\arraybackslash}Xrrrrr@{}}
\toprule
Backend & R@1 & R@2 & R@5 & R@10 & AE@10 \\
\midrule
\multicolumn{6}{@{}l}{\itshape LLM decision models} \\
Jev & 34.30 & 62.18 & 80.68 & 86.62 & 69.50 \\
d1-ultra & 37.57 & 68.93 & 86.12 & 89.72 & 76.20 \\
elephant-1 & 34.91 & 63.23 & 79.68 & 84.38 & 66.60 \\
KnowLine-4B & 32.67 & 57.73 & 75.98 & 81.73 & 61.80 \\
KnowLine-35B-A3B & 32.58 & 60.85 & 81.14 & 86.80 & 70.10 \\
\multicolumn{6}{@{}l}{\itshape Rerankers} \\
qwen3-rerank & 22.64 & 34.75 & 55.55 & 71.61 & 44.60 \\
Qwen3-Reranker-0.6B & 25.67 & 40.81 & 62.54 & 74.18 & 48.40 \\
Qwen3-Reranker-8B & 33.30 & 56.11 & 73.66 & 81.27 & 59.50 \\
jina-reranker-v3.5 & 31.03 & 50.38 & 65.61 & 72.13 & 43.80 \\
\multicolumn{6}{@{}l}{\itshape Embedding similarity} \\
Qwen3-Embedding-8B & 32.71 & 46.68 & 60.87 & 72.48 & -- \\
BGE-Large-EN-v1.5 & 28.73 & 41.22 & 57.12 & 66.31 & 36.40 \\
\multicolumn{6}{@{}l}{\itshape Generative LLM (Qwen3.6-35B-A3B)} \\
Self-reported probability & 32.30 & 45.38 & 58.32 & 64.52 & 37.10 \\
Output rank & 33.38 & 56.49 & 72.90 & 76.90 & 53.40 \\
\bottomrule
\end{tabularx}
\caption{Scoring backends within the same SAG2 procedure on the 1,000 MuSiQue questions (\%). Lexical retrieval alone reaches 54.50 R@10 and 17.50 AE@10 (Table~\ref{tab:expansion}).}\label{tab:backends}
\end{table}

\paragraph{Additional decision backends.} Table~\ref{tab:backends} lists the alternative LLM decision models shown in Figure~\ref{fig:dm}. The strongest of them, d1-ultra \citep{liquidai2026d1ultra}, is a hosted decision model from Liquid AI that, like Jev, returns probabilities instead of generated text and bills only input tokens. The 9B decision model \mbox{elephant-1} \citep{baselabs2026elephant1} is fine-tuned from \mbox{Qwen3.5-9B}, and KnowLine-4B \citep{pelaai2026knowline4b} and KnowLine-35B-A3B \citep{pelaai2026knowline35b} are decision models fine-tuned from Qwen3.5-4B and Qwen3.5-35B-A3B, respectively, the latter a mixture-of-experts model with 35B total parameters. The weights of all three are public on Hugging Face, those of \mbox{elephant-1} under the MIT license and those of the two KnowLine models under Apache-2.0, and all three run on local hardware. Among the rerankers, Jina Reranker v3.5 and Qwen3-Reranker-0.6B publish their weights under the CC BY-NC 4.0 and Apache-2.0 licenses \citep{jinaai2026rerankerv35,qwen2025reranker06b}.

\paragraph{Rerankers.} Every reranker exceeds BM25 in R@5 and R@10 (Table~\ref{tab:backends}), so the seed selection and term expansion of SAG2 raise recall with any relevance scorer, and the LLM decision models add the chain-level gain in AE@10 described above.

\paragraph{Embedding similarity as the scorer.} With the decision-model score replaced by the embedding similarity between the question and each candidate, and the rest of the procedure unchanged, Qwen3-Embedding-8B stays within 1.59 points of Jev at R@1 but falls 14.14 to 19.81 points behind at R@2 to R@10. BGE-Large-EN-v1.5 trails at every cutoff (Table~\ref{tab:backends}). Similarity to the question finds the first-hop passage, which repeats the question's wording, but a later-hop passage resembles its seed rather than the question (Section~\ref{sec:hypergraph}), and no context enters a similarity score, so the deficit begins at the second passage.

\paragraph{Generative LLM as the scorer.} Qwen3.6-35B-A3B, an FP8 build served locally through a chat interface, replaces the decision model under two output protocols. Each scoring round sends the question, the shared context, and all candidates in one request, decoded at temperature 0. The model either returns a JSON object with a probability in \([0,1]\) for each candidate identifier or returns the candidate identifiers as an array ordered from best to worst, where position \(i\) of \(m\) scores \((m-i)/m\). Both protocols use about as many judgments as the default run (326,510 and 327,215). Table~\ref{tab:backends} gives their R@10 and AE@10 beside those of KnowLine-35B-A3B, the decision model fine-tuned from Qwen3.5-35B-A3B, the preceding release of the same architecture, and Section~\ref{sec:analysis} draws the comparison.

\paragraph{Rerankers on three datasets.} With Jina Reranker v3.5 \citep{jinaai2026rerankerv35} in place of Jev and the SAG2 retrieval policy unchanged, SAG2 stays above BM25 on all three datasets, by 17.63, 15.37, and 10.05 R@10 points (Table~\ref{tab:rerank-substitution}). A reranker scores pairs rather than chains, yet the margin over BM25 holds on every dataset because the seeds and the term expansion of Section~\ref{sec:hypergraph} act on whatever scores they receive: the scorer sets the size of the gain, the retrieval policy its presence.

\begin{table}[t]\centering
\footnotesize\setlength{\tabcolsep}{4pt}
\begin{tabular}{@{}lrrr@{}}
\toprule
Method & MuSiQue & 2WikiMultiHopQA & HotpotQA \\
\midrule
SAG2 (Jina) & 72.13 & 86.95 & 97.00 \\
\bottomrule\end{tabular}\caption{R@10 (\%) of SAG2 with jina-reranker-v3.5 as the scorer in place of the decision model on the three 1,000-question subsets. BM25 and SAG2 with Jev are in Table~\ref{tab:main-results}.}\label{tab:rerank-substitution}
\end{table}

\section{Agentic SAG2}\label{app:agentic}
SAG2 can also serve as a retrieval tool for a generative model. In Agentic SAG2, a large language model calls SAG2 with the original question, reads the returned source passages, records which relations are confirmed and which remain unresolved, and decides whether to search again or stop. After its last call, it selects and ranks ten passages from the candidates returned by the tool. Each tool call runs one round of SAG2 expansion. Table~\ref{tab:agentic} reports the results, and the procedure, configuration (Table~\ref{tab:params}), and usage follow below.

The model calls SAG2 2.34 times per question on MuSiQue, where 60.5\% of questions use all three calls, but only 1.47 and 1.33 times on 2WikiMultiHopQA and HotpotQA, where 70.8\% and 79.8\% of questions finish after one call (Table~\ref{tab:agentic-full}).

\textbf{Procedure.} Agentic SAG2 uses the same SQLite FTS5 index and SAG2 retrieval implementation as SAG2. Each question is processed as follows:
\begin{enumerate}
\tightlist
\item The language model receives the original question and the retrieval rules, and makes its first SAG2 tool call with the original question text.
\item SAG2 runs BM25, then the configured number of term-expansion rounds with decision-model scoring, and returns candidate passages with their identifiers.
\item The model reads the results in the full conversation history, records confirmed and unresolved relations, and either calls the tool again for a missing relation or stops.
\item The model outputs a ranking of candidate identifiers, which are mapped to source passages and deduplicated before scoring.
\end{enumerate}
The model decides planning, stopping, and the final ranking, without a separate planner or verifier. Each call carries the original question, the evidence found so far, and the unchanged decision-model instruction; later calls may add query text written by the model. Table~\ref{tab:params} lists the evaluation configuration. In the run with Qwen3.6-35B-A3B as the agent (Table~\ref{tab:agentic}), the open-weight model is served locally through its chat interface with thinking disabled, drives the same tool under the same budget, and makes 2.65 SAG2 calls per MuSiQue question. Its gain over SAG2 in R@10 grows with the number of hops, from 2.89 points on two-hop questions to 6.75 on three-hop and 8.28 on four-hop questions, and it uses 1.68 times as many judgments per question. A later call is issued for a relation the earlier evidence left unresolved and expands from query text that names it, whereas SAG2's two rounds expand from whichever seeds score highest (Section~\ref{sec:hypergraph}). The longest chains leave the most relations unresolved after two rounds, and the gain concentrates there. Agentic BM25 and Agentic BGE dense (Table~\ref{tab:agentic}) keep this agent, budget, and procedure and replace the tool with BM25 retrieval or BGE-Large-EN-v1.5 dense retrieval over the same passages, without decision-model scoring or term expansion.

\begin{table}[t]
\centering\footnotesize\setlength{\tabcolsep}{4pt}
\begin{tabular}{@{}lrrr@{}}
\toprule
& \multicolumn{3}{c}{Tool calls} \\
\cmidrule(l){2-4}
Dataset & 1 & 2 & 3 \\
\midrule
MuSiQue & 268 & 127 & 605 \\
MuSiQue (Qwen3.6-35B-A3B) & 58 & 227 & 712 \\
2WikiMultiHopQA & 708 & 115 & 177 \\
HotpotQA & 798 & 73 & 129 \\
\bottomrule
\end{tabular}
\caption{Number of questions by tool calls in Agentic SAG2, 1,000 questions per dataset; rows without a model name use gpt-6.1-sol. With Qwen3.6-35B-A3B, three questions make no call.}\label{tab:agentic-full}
\end{table}

\textbf{Per-question comparison.} All gains of Agentic SAG2 over SAG2 in Table~\ref{tab:agentic} are significant at the 95\% level under a paired percentile bootstrap over questions; R@10 is higher on 233, 68, and 15 questions and lower on 23, 2, and 2 on MuSiQue, 2WikiMultiHopQA, and HotpotQA.

\section{Extended Related Work}\label{app:related}
This appendix extends Section~\ref{sec:related}.

\paragraph{Reranking and decision models.} Neural rerankers make finer relevance decisions over BM25 or dense candidates \citep{nogueira2019passage,nogueira2020monot5}, and LLM rerankers order a candidate list listwise or setwise \citep{sun2023rankgpt,zhuang2024setwise}; the attainable recall of both is bounded by the pool. LLMs have also served as relevance assessors \citep{thomas2024llmpreferences} and as judges of a passage's utility for answering \citep{zhang2024utility}, and Jev has been evaluated for relevance assessment and reranking of fixed BM25 candidates in Brazilian Portuguese \citep{fernandes2026jevjudge}. In all of these, the model judges a fixed set; SAG2 places the decision model's scores inside the retrieval loop, where they select the seeds and scale the expansion priorities of the next round.

\paragraph{Feedback and adaptive expansion.} Relevance models derive query terms from retrieved documents \citep{lavrenko2001relevance}, RM3 interpolates such a model with the original query \citep{abduljaleel2004rm3}, and LLM-assisted pseudo-relevance feedback computes RM3 only over the top-ranked documents an LLM accepts as relevant \citep{otero2026llmprf}. Feedback text has also been chosen by neural judgments in BERT-QE \citep{zheng2020bertqe} or by an LLM in CSQE \citep{lei2024csqe}. Other LLM expansion generates text for the query: a pseudo-document in query2doc \citep{wang2023query2doc}, hypothetical documents embedded for retrieval in HyDE \citep{gao2023hyde}, answers generated from a prompt that contains the query's BM25 candidates in LameR \citep{shen2024lamer}, and pseudo-passages conditioned on graded LLM relevance judgments from earlier rounds in Seek \citep{bigdeli2026seek}. EfficientRAG trains small token classifiers that tag each chunk as continuing or terminating the search and compose the next-hop query from useful tokens \citep{wang2024efficientrag}. GAR expands the reranking pool through a precomputed corpus graph \citep{macavaney2022gar}, SlideGar extends this to LLM listwise rerankers \citep{rathee2025slidegar}, and GAREN lets a navigator choose which LLM-summarized groups of an offline \(k\)-NN corpus graph to expand \citep{park2026garen}; our RM3 and GAR-BM25 controls (Appendix~\ref{app:controls}) keep these mechanisms without decision-model feedback inside the loop. SAG2 combines three properties that none of these methods has together: one answer-chain judgment in a shared context both selects the feedback passages and scales the expansion priorities of their terms, each activated term retrieves its own list of unvisited passages instead of contributing to one expanded query, and candidates are rescored over several rounds.

\paragraph{Graphs built from lexical statistics or recognized entities.} Several graph-based RAG systems reduce LLM extraction at indexing time: LazyGraphRAG builds a noun-phrase co-occurrence graph and defers LLM use to query time \citep{edge2024lazygraphrag}, TIGRAG a token co-occurrence graph \citep{bonifazi2026tigrag}, LinearRAG and EHRAG an entity graph and an entity hypergraph from named-entity recognition \citep{zhuang2025linearrag,song2026ehrag}, HyperSU hyperedges from entities recognized by an encoder-based extractor \citep{liu2026hypersu}, and KET-RAG a keyword--chunk bipartite graph beside an LLM-extracted skeleton \citep{huang2025ketrag}; KGP lets an LLM agent traverse a passage graph built from lexical or semantic similarity \citep{wang2024kgp}. These systems still construct a graph before retrieval, and most also compute embeddings offline. Shared words link text units in other graph models: TF--IDF supplies the word--sentence edge features of HeterSumGraph \citep{wang2020hetersumgraph}, TF--IDF or BM25 selects the keyword nodes of keyphrase-generation graphs \citep{ye2021keyphrasegraph,kim2021structurekeyphrase}, and ClassKG reselects the keywords of each class by a TF--IDF-style score in each round \citep{zhang2021keywordgraph}. Graphs over retrieved documents have also reranked the candidates of a single retrieval: OkaSim ranks BM25-filtered documents by SimRank in a document--term bipartite graph \citep{champclaux2009okasim}, structural re-ranking links retrieved documents by language-model generation probabilities and re-ranks them by centrality \citep{kurland2005pagerank}, and GRADA propagates scores over symmetrized BM25 links to demote adversarial documents \citep{zheng2025grada}. In these rerankers the ranking stays within a fixed candidate set, whereas SAG2 retrieves unvisited passages beyond the candidate pool.

\paragraph{Hypergraphs and multi-hop retrieval.} Hypergraph-of-entity indexes each document as a hyperedge over its terms and entities and ranks documents by random walks \citep{devezas2019hypergraphofentity}, \citet{spitz2020hypergraph} compare implicit and materialized hyperedges of document collections, and Graph-of-word stores only graph-derived term weights in the inverted index \citep{rousseau2013graphofword}; SAG2 likewise reads passages as hyperedges over terms from the inverted index (Section~\ref{sec:hypergraph}). Multi-hop retrievers condition each step on earlier evidence through next-hop entities that a reader extracts \citep{ding2019cogqa,das2019multistep}, a dense retriever conditioned on retrieved passages \citep{xiong2021mdr}, condensed facts \citep{khattab2021baleen}, or a beam of partial passage chains \citep{zhang2024beamretrieval}, or they express the next hop as text: a generated query per hop \citep{qi2019golden}, reasoning interleaved with BM25 retrieval \citep{trivedi2023ircot}, the previous round's output \citep{shao2023iterretgen}, LLM-generated search phrases matched by word overlap \citep{wang2025elite}, an LLM judge conditioned on a retrieved bridge passage \citep{bacellar2026bridgerag}, or reasoning plans as feedback for retrieving bridge documents \citep{acl2026repair}. PropRAG extracts propositions offline with an LLM and discovers reasoning chains by an LLM-free beam search over proposition paths in a prebuilt graph \citep{wang2025proprag}. LLM agents issue queries as they reason, when prompted \citep{yao2023react,press2023selfask,li2025searcho1} or after reinforcement learning \citep{jin2025searchr1}, search a raw corpus with shell tools \citep{li2026dci}, or explore a workspace that BM25 fills with their results \citep{zhuang2026rise}. SAG2 selects next-round terms from passages the decision model has judged and retrieves with them from a lexical index, without generating query text, training a reader or encoder, or computing embeddings.

\end{document}